\documentclass[sigconf,nonacm]{acmart}
\setcopyright{none}
\renewcommand\footnotetextcopyrightpermission[1]{}

\usepackage{amsmath,amssymb}
\usepackage{graphicx,xspace}
\usepackage{booktabs,tabularx}
\usepackage{tikz}
\usepackage{pifont}
\usepackage{hyperref}
\usepackage{colortbl}
\usepackage{caption}
\usepackage{dblfloatfix}
\usepackage{xurl} % Allows long URLs to break across lines
\DeclareRobustCommand{\circlednum}[1]{%
    \tikz[baseline=(char.base)]{
        \node[
            shape=circle,
            fill=black,
            text=white,
            inner sep=1pt
        ] (char) {\fontsize{7pt}{7pt}\selectfont #1};
    }%
}
\usetikzlibrary{arrows.meta,positioning,fit}

\newcommand{\sys}{\texttt{EmbodiedIntentBench}\xspace}
\begin{document}
\title{\sys: Benchmarking Intent-Driven Runtime Adaptation for Mobile Robotic Missions}
\author{Rajat Bhattacharjya}
\authornote{Correspondence: \texttt{rajatb1@uci.edu}.\\ Paper is currently under review. Authors' version posted for personal use and not for redistribution. Copyright may be transferred without notice, after which this version may
no longer be accessible.}
\affiliation{\institution{University of California, Irvine}\country{USA}}

\author{Changwoon Lee}
\affiliation{\institution{Yonsei University}\country{South Korea}}

\author{Hyeonjong Ju}
\affiliation{\institution{Yonsei University}\country{South Korea}}

\author{Sing-Yao Wu}
\affiliation{\institution{University of California, Irvine}\country{USA}}

\author{Hamidreza Alikhani}
\affiliation{\institution{University of California, Irvine}\country{USA}}

\author{Zhuoyi Yang}
\affiliation{\institution{University of California, Irvine}\country{USA}}

\author{Ian Harris}
\affiliation{\institution{University of California, Irvine}\country{USA}}

\author{Nalini Venkatasubramanian}
\affiliation{\institution{University of California, Irvine}\country{USA}}

\author{Nikil Dutt}
\affiliation{\institution{University of California, Irvine}\country{USA}}

% Compact first-page author layout for the arXiv version.
\makeatletter
\def\@mkauthors{%
  \gdef\@currentauthors{}%
  \gdef\@currentaffiliation{}%
  \global\setbox\mktitle@bx=\vbox{%
    \unvbox\mktitle@bx
    \centering
    {\large
    Rajat Bhattacharjya\textsuperscript{1,*}, Changwoon Lee\textsuperscript{2}, Hyeonjong Ju\textsuperscript{2}, Sing-Yao Wu\textsuperscript{1},
    Hamidreza Alikhani\textsuperscript{1}, Zhuoyi Yang\textsuperscript{1},\\ Ian Harris\textsuperscript{1}, Nalini Venkatasubramanian\textsuperscript{1}, Nikil Dutt\textsuperscript{1}\par}
    \vspace{3pt}
    { \textsuperscript{1}University of California, Irvine, USA \quad \textsuperscript{2}Yonsei University, South Korea\par}
    \medskip
  }%
}
\makeatother
\renewcommand{\shortauthors}{Bhattacharjya et al.}
\begin{abstract}
Mobile robotic sensing systems must adapt to operator requirements, limited resources, changing connectivity, and observation opportunities shaped by physical motion. Evaluating runtime strategies therefore requires assessing whether their decisions fulfill mission objectives, beyond isolated perception accuracy or execution efficiency. We present \sys, a benchmark for intent-driven runtime adaptation in mobile robotic missions. It represents mission intent through structured contracts for coverage, temporal evidence, reporting, urgency, and resources, and couples simulated unmanned aerial vehicle (UAV) workloads with causal, resource-accounted execution. Computation decisions affect mission time, resource use, and which observations can contribute evidence. We evaluate fixed and adaptive perception strategies in multi-phase missions and under changing urgency, connectivity, and battery conditions. In the multi-phase Depot study, a configuration with 87.5\% full-bank target recall across visible target observations completes only 1 of 16 missions, whereas a lower-recall configuration completes 14. In an ablation, ignoring execution-induced observation loss increases aggregate mission success and reverses the ordering of two runtime strategies. Replays using Jetson AGX Orin CPU and GPU profiles change the relative success of local and remote strategies. In selected Jetson GPU-profile replays, intent-driven adaptation reduces modeled task energy by up to 28\% with unchanged mission success. \sys provides an extensible framework for evaluating runtime decisions against mission fulfillment under physical and resource constraints.
\end{abstract}
\keywords{embodied robotic missions, intent-driven adaptation,
runtime computing systems, mission-level benchmarking,
cyber-physical systems}
\maketitle

%\vspace{-3ex}
\section{Introduction}
\label{sec:intro}

Mobile robotic systems, particularly unmanned aerial vehicles (UAVs), are increasingly employed in mission-critical applications such as search and rescue, disaster response, and infrastructure inspection~\cite{bhattacharjya2025avery,cladera2025air,farahani2026evlm}. Advances in language-guided robotics and intent-driven embodied intelligence are enabling operators to specify \emph{what} a mission must accomplish, rather than prescribing \emph{how} individual sensing and computation tasks should be executed~\cite{cladera2025air,bhattacharjya2025avery,farahani2026evlm, 10.1145/3715014.3722081}. However, translating such high-level natural-language intent into effective runtime behavior remains challenging under constrained onboard computation, limited energy, and variable network connectivity~\cite{bhattacharjya2025avery}.

\begin{figure}[t]
    \centering
    \includegraphics[width=\columnwidth]{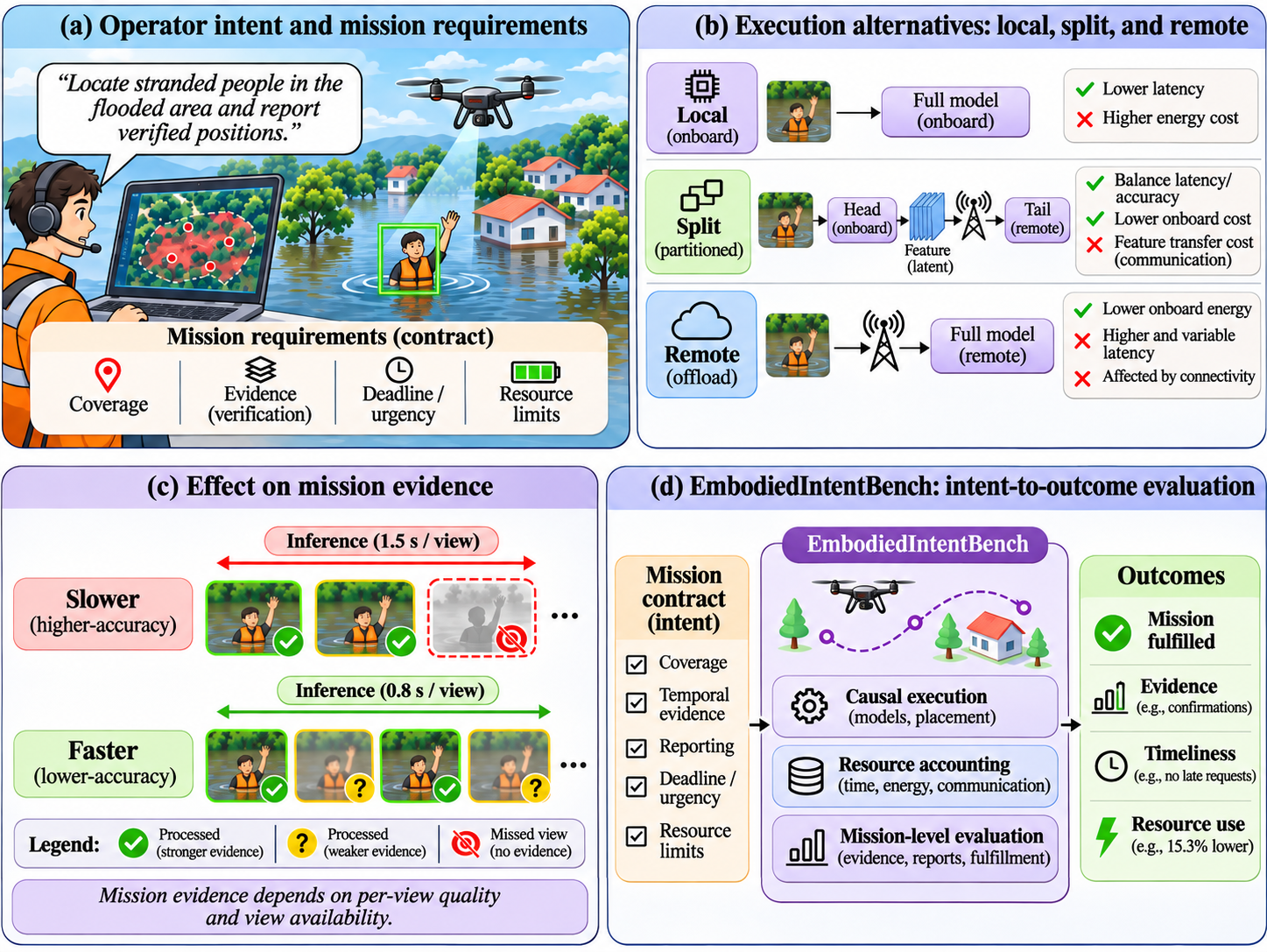}
    %\vspace{-4pt}
    \caption{\textbf{Motivation and overview of \sys.}
    (a) Operator-specified missions impose requirements
    for target coverage, evidence verification,
    timeliness, and resource use.
    (b) Local, split, and remote execution offer
    different computation and communication tradeoffs;
    the current evaluation considers whole-model
    local and remote execution.
    (c) Perception quality and execution latency
    influence mission evidence, as longer processing
    times can prevent subsequent observations
    from being processed.
    (d) \sys connects structured mission requirements
    with physically situated, resource-accounted
    execution and mission-level evaluation.}
    \label{fig:motivation}
    %\vspace{-5ex}
%\vspace{-8pt}
\end{figure}

Consider a UAV deployed to locate stranded individuals following a flood, where an operator requests verified locations to support rescue operations (Fig.~\ref{fig:motivation}(a)). An initial search may prioritize broad target coverage, while subsequent rescue operations may require corroborated findings or more timely updates. Although these requests concern the same environment, they impose different evidence, timing, and resource requirements~\cite{bhattacharjya2026floodreasonbench}. Mission intent describes an operator's desired outcome and is represented through a structured contract, which can be supplied directly or constructed from a supported natural-language request.

Fulfilling such requirements cannot be assessed through perception accuracy or execution efficiency alone. A correct prediction may arrive too late, while a timely detection may lack sufficient evidence for confirmation. Even with sufficient evidence, reporting errors or resource violations can prevent mission completion. Consequently, successful individual computations do not necessarily translate into successful mission execution~\cite{weng2026multi, chen2026controller}.

These distinctions become particularly important for mobile robotic systems, where physical motion creates time-sensitive observation opportunities. As a UAV traverses its environment, useful views emerge and disappear with changing viewpoints, occlusions, and target visibility. A computationally demanding perception configuration may improve an individual prediction while preventing subsequent observations from being processed (Fig.~\ref{fig:motivation}(c))~\cite{soyyigit2025mural, 10.1145/3774906.3800471}. Such tradeoffs also depend on whether computation executes onboard or remotely (Fig.~\ref{fig:motivation}(b)), with different implications for latency, communication, and energy consumption~\cite{kim2026approxbit, azimi2026cloud, candal2023drone}. Runtime decisions must therefore be evaluated through their contribution to mission fulfillment, rather than isolated measures of prediction quality or execution efficiency.

Prior work has established the importance of evaluating robotic computation in the context of physical execution and downstream task performance~\cite{bala2024ooda}. MAVBench~\cite{boroujerdian2018mavbench} characterizes interactions among computation, physical dynamics, and resource consumption, while RobotPerf~\cite{mayoral2024robotperf} and A2Perf~\cite{uchendu2025a2perf} provide robotic computing benchmarks. Streaming perception~\cite{li2020towards, yang2022streaming} and control-aware prediction~\cite{mcallister2022control} highlight the importance of prediction timing and downstream consequences. More recently, DEEP~\cite{cong2026decision} and LIBERO-Dynamic~\cite{ren2026libero} examine decision-level computation value and inference latency. However, these approaches do not jointly evaluate runtime computation against explicit mission requirements involving accumulated evidence, temporal verification, reporting, and resource constraints.

This motivates a fundamental question: \textbf{\emph{How can runtime sensing and computation strategies be systematically evaluated based on their ability to fulfill operator-specified mission requirements under changing physical and resource conditions?}} Addressing this requires explicit mission requirements, causal accounting of computation and observation opportunities, and evaluation of whether accumulated evidence and reports satisfy the intended mission.

To this end, we present \textbf{\sys}, a benchmark for evaluating intent-driven runtime adaptation in mobile robotic missions (Fig.~\ref{fig:motivation}(d)). \sys operationalizes mission intent through structured \emph{mission contracts} specifying target coverage, temporal evidence, verification, reporting, urgency, and resource constraints. {It combines Gazebo-based UAV workloads~\cite{koenig2004design} with causal, resource-accounted execution; using shared recorded trajectories, it measures how runtime computation changes observation availability, resource use, and mission fulfillment.} The current evaluation isolates computation decisions over fixed recorded UAV trajectories; navigation and physical control are outside its scope.

We demonstrate \sys through two complementary case studies. The first examines perception choices in multi-phase Lake and Depot missions (lake and warehouse) involving coverage, repeated confirmation, timely reporting, and fresh revisits. The second investigates runtime adaptation under changing urgency, connectivity, and battery conditions using whole-model local and remote execution. {We also examine how mission requirements can guide when perception stops, using Jetson-derived profiles and measuring energy for selected schedules on the device.} Our evaluation reveals that a configuration achieving 87.5\% full-bank target recall (across visible target observations) completes only 1 of 16 missions, whereas another with lower recall completes 14. Further validation shows that ignoring execution-induced observation loss can overestimate mission success and reverse the relative ranking of runtime strategies.

The main contributions of this paper are as follows:
%\vspace{-2ex}
\begin{itemize}
    \item \textbf{Intent-driven mission specification.}
    We introduce structured mission contracts expressing operator objectives through explicit requirements for target coverage, temporal evidence, verification, reporting, urgency, and resource constraints.

    \item \textbf{Causal, resource-accounted execution.}
    We develop an evaluation methodology coupling runtime computation with physically situated observation opportunities, accounting for how processing delays, communication, and onboard resource consumption affect mission evidence.

    \item \textbf{Mission-level evaluation and diagnostics.}
    We connect runtime decisions to evidence acquisition, confirmation, reporting, and mission fulfillment, enabling analysis of the requirements and execution conditions underlying mission outcomes.

    \item \textbf{Extensible benchmark and empirical characterization.}
    We provide configurable workloads, reference strategies, and extension interfaces, demonstrating discrepancies between perception performance and mission success, the effects of causal execution, and the consequences of changing resource conditions.\footnote{{A public artifact provides code, inputs, and reference results for the dynamic study and the Jetson and intent-driven analyses; Appendix A describes its scope. The complete benchmark artifact, including the multi-phase workloads and prediction banks, will be released upon acceptance. Artifact: \url{https://github.com/rajatbh21/EmbodiedIntentBench-artifact}}}
\end{itemize}

% The remainder of this paper is organized as follows. Sec.~\ref{sec:design} presents the benchmark architecture, followed by its implementation and extension interfaces in Sec.~\ref{sec:implementation}. Sec.~\ref{sec:experimental_methodology} describes the experimental methodology, and Sec.~\ref{sec:results} presents the evaluation and validation studies. Related work and conclusions follow in Sec.~\ref{sec:related_work} and Sec.~\ref{sec:conclusion}, respectively.

\vspace{-9pt}
\section{\sys: Design and Architecture}
\label{sec:design}

\begin{figure*}[t]
    \centering
    \includegraphics[width=0.9\textwidth]{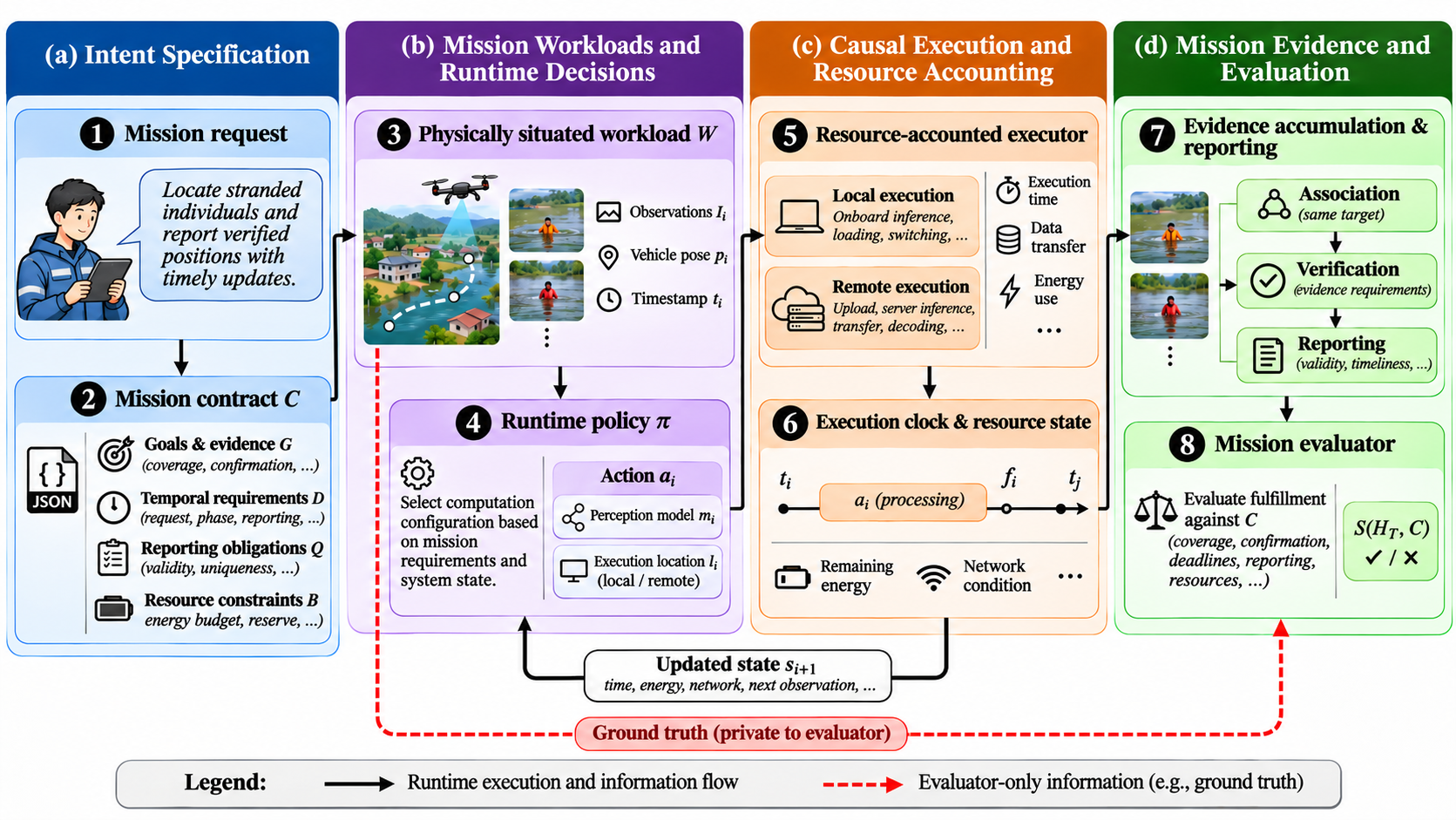}

\vspace{-4pt}

\caption{\textbf{Architecture of \sys.}
(a) \emph{Intent specification:} An operator request
(\circlednum{1}) is represented through a structured
mission contract (\circlednum{2}) defining evidence,
temporal, reporting, and resource requirements.
(b) \emph{Mission workloads and runtime decisions:}
Physically situated observations (\circlednum{3})
and current system state inform a runtime policy
(\circlednum{4}) that selects a computation configuration.
(c) \emph{Causal execution and resource accounting:}
A resource-accounted executor (\circlednum{5})
applies the selected computation and its associated
costs, while the execution clock and resource state
(\circlednum{6}) determine result availability and
subsequent observation opportunities.
(d) \emph{Mission evidence and evaluation:}
Admitted results contribute to evidence accumulation
and reporting (\circlednum{7}), while a mission evaluator
(\circlednum{8}) assesses fulfillment against the contract.
Solid arrows indicate runtime execution and information
flow, including state feedback; red dashed arrows
indicate private ground truth accessible only
to the evaluator.}

\label{fig:architecture}

\vspace{-8pt}
\end{figure*}

\vspace{-4pt}
\subsection{Architecture Overview}
\label{sec:architecture_overview}

% \sys connects operator-specified mission requirements
% to runtime computation, physical execution, and
% mission-level outcomes. Fig.~\ref{fig:architecture}
% illustrates the architecture, organized into four
% interconnected components.

% First, an operator's mission request
% (\circlednum{1}) is represented through a structured
% \emph{mission contract} (\circlednum{2}), defining
% the evidence, timing, reporting, and resource
% requirements of the mission. Second, mobile robotic
% workloads provide time-varying observation opportunities
% (\circlednum{3}), while a runtime policy
% (\circlednum{4}) selects computation configurations
% according to the mission requirements and current
% system state. Third, a resource-accounted executor
% (\circlednum{5}) applies these decisions, while
% the execution clock and resource state
% (\circlednum{6}) determine when results become
% available and which subsequent observations can
% be processed. Finally, admitted outputs contribute
% to mission evidence and reports (\circlednum{7}),
% which are evaluated against the mission contract
% (\circlednum{8}).

% The architecture separates mission specification,
% runtime decision-making, and evaluation. Competing
% strategies operate under common mission and
% environmental conditions, while their execution
% decisions determine the computation performed,
% its costs, and the evidence ultimately available.
% This allows \sys to evaluate different runtime
% strategies without prescribing a particular
% perception model, computation arrangement,
% or adaptation algorithm.

\sys connects operator-specified mission
requirements to runtime computation and
mission-level outcomes, as illustrated in
Fig.~\ref{fig:architecture}. A structured
mission contract defines successful execution,
while a runtime policy selects computation
configurations based on mission requirements
and current system state. Causal, resource-accounted
execution determines result availability,
resource consumption, and subsequent observation
opportunities. Admitted outputs contribute to
mission evidence and reporting, which are
evaluated against the contract. This separation
enables controlled comparisons of runtime
strategies without prescribing a particular
perception model or adaptation algorithm.

\vspace{-9pt}
\subsection{Intent Specification and Mission Contracts}
\label{sec:mission_contract}

\noindent\textbf{Intent specification
(\circlednum{1}--\circlednum{2}).}
% Previous wording marked for deletion: 
The first component of \sys establishes what constitutes successful mission execution.
% \textcolor{blue}{This component defines successful mission execution.}
As illustrated in Fig.~\ref{fig:architecture}(a),
an operator-facing request is represented through
a structured \emph{mission contract} that makes
the intended sensing objectives and execution
requirements explicit. For example, a request
to locate stranded individuals and report
verified positions can specify the required
target coverage, corroborating observations,
and reporting constraints.

The contract provides a common interface
between operator intent, runtime policies,
and mission evaluation. Requirements are
expressed through a validated JSON-based
specification, which can be supplied directly
or constructed from supported operator-facing
descriptions. The benchmark operates on the
structured requirements rather than assuming
unrestricted natural-language interpretation
during execution.

We represent a mission contract as
\begin{equation}
    \mathcal{C} =
    (\mathcal{G},\mathcal{D},\mathcal{Q},\mathcal{B}),
    \label{eq:mission_contract}
\end{equation}
where $\mathcal{G}$ defines mission goals and
evidence requirements, $\mathcal{D}$ specifies
temporal constraints, $\mathcal{Q}$ captures
reporting obligations, and $\mathcal{B}$
represents resource constraints.

\noindent\emph{Mission goals and evidence
($\mathcal{G}$).}
The mission specification identifies the
targets or regions of interest and the
evidence required for their successful
identification. Requirements may include
target coverage, repeated observations,
minimum temporal separation between
confirmations, and evidence collected
during distinct mission phases. For
multi-phase missions, a later phase may
require fresh observations, preventing
previously collected evidence from
automatically satisfying a new requirement.

\noindent\emph{Temporal requirements
($\mathcal{D}$).}
Temporal constraints operate at different
levels, including individual perception
requests, mission phases, and final reporting.
For a perception request released at time
$r_i$, an execution allowance $\Delta_i$
defines its deadline:
\begin{equation}
    d_i = r_i + \Delta_i.
    \label{eq:request_deadline}
\end{equation}
Urgency may evolve during execution,
changing the allowance assigned to newly
released requests without necessarily
changing the underlying sensing objective.
Phase and reporting deadlines independently
constrain when the required evidence and
findings must become available.

\noindent\emph{Reporting obligations
($\mathcal{Q}$).}
Reporting requirements specify which findings
must be delivered, their temporal validity,
and admissible limits on incorrect or
duplicate reports. These requirements
distinguish identifying a target in an
observation from producing a valid mission
report.

\noindent\emph{Resource constraints
($\mathcal{B}$).}
The contract defines applicable resource
limits, such as onboard energy budgets,
required battery reserves, and communication
constraints. These limits are distinct from
the instantaneous resource state observed
during execution.

The mission contract thus defines both
the requirements exposed to a runtime
policy and the criteria used to evaluate
its execution. However, the actions
available to a policy also depend on
the observations and operating conditions
encountered during the mission. 

\vspace{-9pt}
\subsection{Mission Workloads and Runtime Decisions}
\label{sec:workloads_policies}

\noindent\textbf{Physically situated workloads
(\circlednum{3}).}
\sys represents mobile robotic missions
through time-indexed observations associated
with physical execution. A workload specifies
the observation opportunities presented to
the runtime, together with the physical
context and evaluation annotations required
to determine mission outcomes. In the current implementation, Gazebo-based
UAV missions~\cite{koenig2004design} provide
recorded trajectories with synchronized
RGB observations, vehicle poses, and
acquisition timestamps. \\A workload is
represented as
\begin{equation}
    \mathcal{W} =
    \{(I_i,\mathbf{p}_i,t_i)\}_{i=1}^{N},
    \label{eq:workload}
\end{equation}
where $I_i$ is the image acquired at time
$t_i$ and $\mathbf{p}_i$ denotes the
corresponding vehicle pose. Ground-truth
annotations are retained separately
for evaluation.

These observations define a common set
of \emph{potential observation opportunities}
across competing strategies. Their actual
processed subsets, however, depend on
the execution decisions made during the
mission. Holding the physical trajectory
fixed within a comparison isolates the
effect of runtime computation on observation
availability without conflating it with
changes in navigation behavior.

To support reproducible comparisons, the
implementation constructs a bank of
predictions by evaluating candidate models
on recorded observations. During execution,
only the prediction associated with the
selected model and observation can enter
downstream processing, subject to its
execution and timing requirements.
Unselected predictions remain unavailable
to the runtime policy.

\noindent\textbf{Runtime policy and computation
selection (\circlednum{4}).}
At each eligible decision point, a runtime
policy selects from the computation
configurations supported by the workload.
Let $s_i$ represent the public operating
state and $\mathcal{A}$ the available
action space. A policy $\pi$ selects
\begin{equation}
    a_i = \pi(I_i,\mathcal{C},s_i)
    \in \mathcal{A},
    \label{eq:runtime_action}
\end{equation}
where the action may specify a perception
model, execution settings, computation
placement, or other supported processing
choices.

The operating state may include current
network conditions, available onboard
resources, request deadlines, and declared
execution profiles. Where supported,
information derived from previously
completed, selected actions may also
inform subsequent decisions. 
Ground-truth target identities and correctness
labels are reserved for the evaluator. Future
observations and outputs of unselected perception
configurations are likewise inaccessible through
the runtime policy's decision interface.

For the evaluated perception-routing configurations,
an execution action is instantiated as
$a_i=(m_i,\ell_i)$, where $m_i$ identifies
a perception configuration and $\ell_i$ denotes
local or remote execution. The action space also
permits abstention, denoted by $a_i=\bot$.
Other execution arrangements, including
partitioned inference, can be incorporated
through configurations with their own processing
stages and execution profiles.

This interface separates the decision
strategy from the consequences of its
actions. A policy may optimize predicted
quality, timeliness, resource feasibility,
mission progress, or a combination of
these objectives. Whether its selected
computation contributes useful evidence,
however, depends on how that computation
executes over time.

\vspace{-9pt}
\subsection{Causal Execution and Resource Accounting}
\label{sec:causal_execution}

\noindent\textbf{Resource-accounted execution
(\circlednum{5}).}
The execution framework applies the time
and resource costs associated with a
selected action, as illustrated in
Fig.~\ref{fig:architecture}(c).
Each execution configuration specifies
the processing operations and associated
costs required to produce its output.

In the current implementation,
\emph{local execution} incurs the
configured onboard inference latency
and energy consumption, together with
applicable preprocessing, model loading,
switching, and supporting operations. \emph{Remote execution} additionally
accounts for input preparation, upload,
server-side processing, response transfer,
and decoding. Communication is evaluated
against the applicable network profile,
including bandwidth changes and outage
intervals. Execution times, transferred
data, and onboard energy are charged
according to the declared execution
and resource profiles.

Importantly, resource accounting is
not restricted to successful inference.
A late computation retains its incurred
processing cost even when its result
cannot contribute timely evidence.
% Similarly, a canceled remote request
% retains the communication and computation
% costs incurred before cancellation.
This ensures that different runtime
strategies are compared against the
resources consumed by their executed
decisions, rather than only the costs
of successful outputs.

\noindent\textbf{Causal execution and observation
availability (\circlednum{6}).}
The benchmark maintains a logical
execution clock that advances with
the operations performed by the runtime.
Execution time determines when a
selected result becomes available and,
under the serial processing rule,
when the next observation can be admitted.

Let $f_i$ denote the completion time
of the blocking processing interval
initiated for observation $i$. The
next eligible observation is
\begin{equation}
    j =
    \min\{k>i \mid t_k \geq f_i\}.
    \label{eq:next_observation}
\end{equation}
If no such observation remains within
the workload, processing reaches the
end of the available observation stream.
Captures acquired during the intervening
interval are recorded as unprocessed
opportunities rather than queued
for later execution. This serial execution model represents blocking perception pipelines, while alternative buffering and asynchronous processing can be incorporated through corresponding execution backends.

Result admission is determined separately
by its applicable timing requirements.
A prediction that becomes available
after its request deadline cannot
contribute as a timely result, although
the work already performed remains
accounted for. Thus, the execution
framework connects the latency of a
selected configuration to both its
immediate result and the subsequent
observation opportunities available
to the mission.

\noindent\textbf{Resource-state evolution.}
In addition to execution time, the
benchmark tracks the resources consumed
throughout a mission. Onboard energy
over a mission horizon $T$ is modeled as
\begin{equation}
\begin{aligned}
E_{\mathrm{onboard}}(T)
={}& \int_0^T P_{\mathrm{flight}}(t)\,dt\\
&+ E_{\mathrm{compute}}
 + E_{\mathrm{comm}}
 + E_{\mathrm{support}},
\end{aligned}
\label{eq:energy_accounting}
\end{equation}
where the terms represent flight,
onboard computation, communication,
and supporting operations, respectively.
Supporting work may include
preprocessing, routing, loading,
decoding, association, and verification,
according to the execution profile.

Costs are charged at their declared
execution boundaries to avoid duplicate
accounting. The recorded flight duration
is charged in full, and the framework
tracks remaining energy, required
reserve, and communication consumption
as execution proceeds. The resulting time and resource state
feeds back into subsequent policy
decisions. Consequently, the selected
action affects not only the current
prediction but also the computation
and observation opportunities available
later in the mission. The outputs that
remain usable are then passed to the
mission evidence pipeline.

\vspace{-9pt}
\subsection{Mission Evidence and Evaluation}
\label{sec:mission_evaluation}

\noindent\textbf{Evidence accumulation and reporting
(\circlednum{7}).}
The evidence pipeline connects
individual perception results to
the requirements of a complete
mission, as illustrated in
Fig.~\ref{fig:architecture}(d).
Admitted predictions are processed
through the applicable downstream
stages, which may include target
association, evidence verification,
and report generation.

Association connects observations
corresponding to the same target. Runtime association uses projected detection
centroids and camera poses, without ground-truth
identities; evaluator matching is performed separately.
Verification determines whether
the accumulated observations satisfy
the mission's evidence requirements,
including repeated confirmations,
temporal separation, and freshness
across phases. Reporting produces
findings subject to the contract's
validity and delivery requirements.

The processing stages and their
requirements depend on the mission
configuration. A preliminary survey
may require a different evidence
threshold from a confirmation or
revisit task, even when the underlying
observations and perception models
are shared. This allows different
mission objectives to be evaluated
through a common execution interface
without imposing a single evidence
criterion on every workload.

\noindent\textbf{Mission-level evaluation
(\circlednum{8}).}
% The evaluator determines whether
% the executed sequence satisfies
% the mission contract. Let
% $\mathcal{H}_T$ denote the history
% of admitted predictions, accumulated
% evidence, generated reports, and
% resource consumption over the
% mission horizon $T$.
The evaluator determines whether the executed
sequence satisfies the mission contract.
Let $\mathcal{H}_T$ denote the mission execution
history over horizon $T$, including selected
actions, execution events, admitted predictions,
accumulated evidence, generated reports, and
resource consumption.

Each contract defines a set of
requirements $\mathcal{R}(\mathcal{C})$.
Complete mission fulfillment is
expressed as
\begin{equation}
S(\mathcal{H}_T,\mathcal{C}) =
\mathbb{I}\left[
\bigwedge_{r\in\mathcal{R}(\mathcal{C})}
r(\mathcal{H}_T)
\right],
\label{eq:mission_success}
\end{equation}
where each requirement evaluates
an applicable condition on the
executed mission history.
These conditions may include
evidence coverage, confirmation,
temporal validity, reporting
correctness and timeliness,
and resource compliance.

Request-level deadlines determine
whether individual outputs are
admitted as timely evidence,
but a missed request does not
necessarily imply complete mission
failure. Mission success depends
on whether the resulting history
satisfies all applicable requirements
of the contract. The evaluator uses private ground
truth to assess prediction
correctness, target identities,
and evidence validity. This
information is excluded from the
runtime policy's decision interface.

Beyond complete mission success,
\sys records diagnostic measures
that characterize progress toward
individual requirements and explain
execution outcomes. These include
target coverage, confirmation,
timely and unique reports,
incorrect or duplicate findings,
request deadline misses,
processed and skipped observations,
and resource consumption.

Together, these measures make
it possible to distinguish whether
a strategy fails because of
insufficient evidence, untimely
execution, reporting deficiencies,
or resource constraints. They
also allow improvements in
individual systems-level metrics
to be examined separately from
their contribution to complete
mission fulfillment. The following section describes how these
architectural components are instantiated
in the current benchmark implementation,
together with the interfaces for extending
its workloads and runtime configurations.

\vspace{-9pt}
\section{\sys: Implementation and Extensibility}
\label{sec:implementation}
\begin{table*}[t]
\centering
\caption{\textbf{Implementation interfaces and
extension mechanisms in \sys.}
Components can be replaced independently,
subject to their input requirements and
corresponding execution or evaluation rules.}
\label{tab:benchmark_interfaces}
\vspace{-4pt}
\footnotesize
\renewcommand{\arraystretch}{1.10}
\begin{tabularx}{\textwidth}{
    @{}p{0.18\textwidth}XX@{}}
\toprule
\textbf{Component} &
\textbf{Required inputs} &
\textbf{Extension mechanism} \\
\midrule

\textbf{Mission specification} &
Structured goals, evidence, timing,
reporting, and resource requirements. &
Add mission contracts and corresponding
evaluation rules. \\

\textbf{Physical workload} &
Timestamped observations, vehicle
poses, and private evaluation labels. &
Import new recordings, environments,
and trajectories. \\

\textbf{Perception configuration} &
Prediction outputs, preprocessing
metadata, and execution profiles. &
Register new models and
configuration-specific costs. \\

\textbf{Runtime policy} &
Public contract, eligible observation,
resource state, and selected history. &
Implement fixed or adaptive
computation-selection strategies. \\

\textbf{Execution backend} &
Selected actions, processing stages,
network conditions, and cost profiles. &
Implement alternative execution
and resource-accounting mechanisms. \\

\textbf{Evidence and evaluation} &
Completed predictions, evidence history,
mission contract, and evaluator truth. &
Replace association and verification
stages or introduce task-specific scorers. \\

\bottomrule
\end{tabularx}
\vspace{-8pt}
\end{table*}
We implement \sys as a configurable benchmark
suite with separate components for workload
preparation, runtime execution, and mission
evaluation. The implementation supports
versioned study configurations, interchangeable
runtime policies and processing components,
and reproducible execution over recorded
sensing workloads.

\vspace{-9pt}
\subsection{Study Assembly and Input Validation}
\label{sec:benchmark_workflow}

A benchmark study combines a structured mission
contract, time-indexed observations, a catalog
of perception configurations, execution and
resource profiles, and a set of runtime policies.
Each study configuration binds these inputs
to the applicable mission evaluator and
operating conditions.

% The implementation separates physical workload
% acquisition from runtime evaluation. Recorded
% RGB images, vehicle poses, and timestamps
% define the potential observation stream.
% Candidate perception configurations are
% evaluated on these observations to construct
% a prediction bank indexed by capture and
% configuration. During replay, this bank is
% accessed only for the selected computation;
% precomputation does not eliminate its
% configured execution costs.

The implementation separates physical workload
acquisition from runtime evaluation. Recorded
RGB images, vehicle poses, and timestamps
define the potential observation stream.
Candidate perception configurations are
evaluated on these observations to construct
a prediction bank indexed by capture and
configuration. During replay, this bank is
accessed only for the selected computation;
precomputation does not eliminate its
configured execution costs.
This separation enables CPU-only benchmark
replay using prepared prediction banks and
execution profiles, without requiring repeated
neural inference or robotic simulation.
GPU acceleration may be used to generate
predictions for new perception configurations,
but is not required for evaluating runtime
strategies over existing workloads.

Study manifests identify the observation
sources, prediction artifacts, model
configurations, mission contracts, and
execution profiles. Input checks verify
the required capture and configuration
coverage, while source and configuration
hashes support consistency checks across
repeated executions. Execution costs retain
their provenance as measurements,
characterized profiles, or declared
assumptions rather than being treated
as equivalent hardware measurements.

Importantly, the runtime receives only
the public projection of the mission
contract, the current eligible observation
and system state, and prediction-derived
history from completed selected actions.
Ground-truth identities, correctness
labels, future observations, and
unselected model outputs remain
inaccessible through the policy interface.
These restrictions provide a controlled
evaluation boundary for trusted policy
implementations.

\vspace{-9pt}
\subsection{Runtime Interfaces and Extension Points}
\label{sec:runtime_interfaces}

\sys exposes interfaces for
mission specification, physical workloads,
perception configurations, runtime policies,
execution backends, and evidence processing.
Table~\ref{tab:benchmark_interfaces}
summarizes their inputs and extension
mechanisms.

The current studies instantiate perception
selection using whole-model local execution
and modeled remote execution. Separate
development experiments also exercise
replaceable association and evidence-
verification stages. These mechanisms
are implemented through defined component
interfaces rather than unrestricted
robotic task execution. Additional arrangements, such as split
inference~\cite{splitsurvey,bhattacharjya2026freshlatent}, require implementations of
their processing stages, intermediate
transfers, and cost profiles. Similarly,
new mission objectives require compatible
evidence-processing and evaluation logic;
arbitrary natural-language mission
interpretation is not assumed.

\vspace{-9pt}
\subsection{Controlled Replay and Reproducibility}
\label{sec:reproducibility}

Each replay initializes fresh policy and
evidence-tracking state under a fixed
recording, mission contract, and operating
condition. Competing policies share the
same potential observations and external
conditions, while their execution decisions
determine which observations are processed
and contribute to mission fulfillment.

Execution traces record selected actions,
prediction accesses, completion times,
skipped observations, resource consumption,
evidence updates, reports, and requirement
outcomes, enabling reconstruction of
individual executions. Validation checks reconstruct execution timing,
resource accounting, and mission outcomes from
recorded inputs, while verifying policy-input
isolation and deterministic regeneration of
evaluation artifacts. These checks establish
replay consistency rather than independently
reproducing Gazebo simulations, neural inference,
or hardware measurements. Thus, repeated
executions provide controlled comparisons over
shared simulated workloads, not independent
robotic deployments.

\vspace{-9pt}
\section{Experimental Methodology}
\label{sec:experimental_methodology}

We evaluate \sys through two complementary studies.
The first examines perception configuration selection
in multi-phase UAV sensing missions, while the second
investigates runtime adaptation under changing request
urgency, connectivity, and battery conditions. We
additionally conduct validation experiments examining
mission-contract sensitivity, execution-dependent
observation availability, mission feasibility, and
failure attribution. All studies use controlled replay
over recorded simulated workloads, enabling comparisons
under matched observation opportunities and operating
conditions.

\vspace{-9pt}
\subsection{Physical Workloads and Mission Requirements}
\label{sec:experimental_workloads}

\begin{figure*}[t]
    \centering
    \includegraphics[width=0.75\textwidth,trim=0bp 26bp 0bp 10bp,clip]{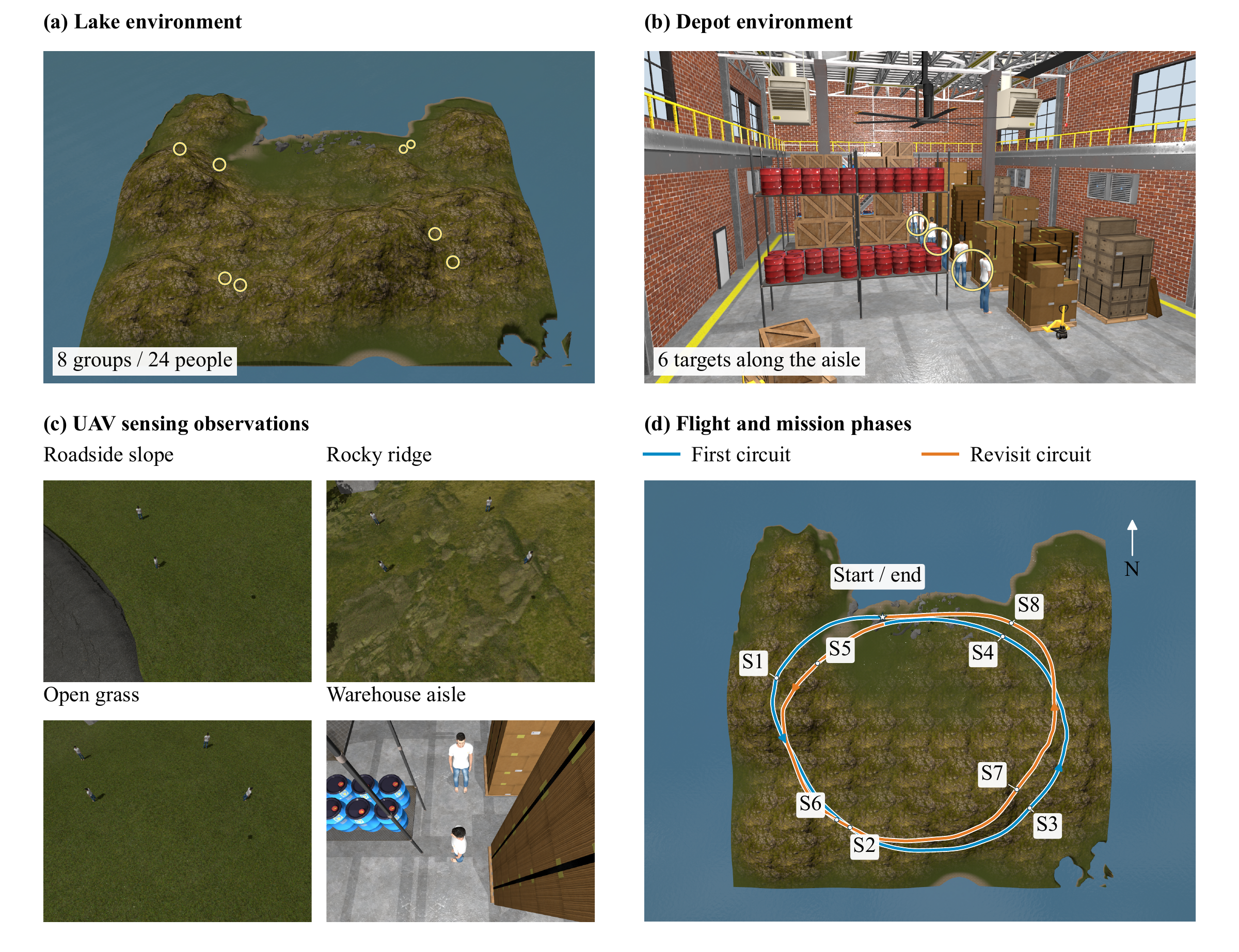}
    % \caption{\textbf{Physical workloads and mission
    % scenarios in \sys.}
    % (a) Lake simulation environment and representative
    % human target locations.
    % (b) Depot environment and representative warehouse
    % observations.
    % (c) Recorded UAV sensing observations illustrating
    % changing target visibility.
    % (d) Representative flight trajectories and
    % multi-phase mission organization, including
    % sector coverage and revisits.
    % Physical trajectories and potential observations
    % remain fixed within each comparison, while runtime
    % decisions determine which observations can be
    % processed and contribute to mission evidence.}
\vspace{-4pt}
\caption{Physical workloads and mission
scenarios in \sys.
(a) Wide-area Lake environment containing
24 people across eight sectors, used for
the dynamic-urgency study.
(b) Depot warehouse environment.
(c) Recorded UAV observations illustrating
variations in target visibility.
(d) Illustrative wide-area sector circuit,
shown for spatial context rather than
as an evaluated trajectory.
The multi-phase Lake study uses separate
six-target corridor layouts, whereas
dynamic-urgency recordings comprise
local passes near focal sectors.
Recorded trajectories remain fixed
within each runtime-policy comparison.}

    \label{fig:workloads}
\vspace{-12pt}
\end{figure*}

\textbf{Multi-phase sensing missions.}
We construct UAV sensing workloads in Gazebo
Harmonic~\cite{koenig2004design,gazeboHarmonic}
using two environments: a Lake scene comprising
terrain, vegetation, water, and residential
structures, and a textured warehouse environment
based on the Depot model. The scenes incorporate
assets from Gazebo Fuel~
\cite{gazeboDepotV6,gazeboHarmonicTerrainV1,
gazeboHarmonicTerrainObjectsV1,gazeboLakeHouseV3,
gazeboCoastWaves2V3}.
Fig.~\ref{fig:workloads} illustrates their
physical settings and observation opportunities.

Each multi-phase scenario designates six static
human targets across three sectors. The Lake
recordings use separate outbound--return corridor
layouts, with all six people designated as targets. The UAV follows a fixed flight trajectory
covering the sectors before revisiting the first.
The Lake and Depot cohorts contain six and four
simulated recordings, respectively. Each recording
provides 60 RGB observations at $960\times720$
resolution, sampled every two seconds over
a 120-second mission.

The mission contracts are motivated by
search-and-rescue requirements for coverage,
verification, reporting, and continued
operation~\cite{natsar2026manual}, while
their numerical thresholds are explicitly
defined benchmark parameters. A rapid-survey
mission requires evidence for at least half
of the designated targets in each phase,
supported by two observations spanning at
least two seconds. A verified-inspection
mission requires all designated targets,
with three observations spanning at least
four seconds.

Both contracts comprise four 30-second
phases, including a revisit requiring fresh
evidence. The respective minimum unique-target
report recall requirements are 0.5 and 0.75,
with reports due by 120 seconds. Each phase
allows at most two false detection components,
while the mission permits at most two
incorrect and two duplicate reports.
An outer execution deadline of 130 seconds
is evaluated separately from the phase
and reporting deadlines.

\textbf{Dynamic-urgency missions.}
\begin{table*}[t]
\centering
\caption{\textbf{Experimental cohorts and
controlled factors.} A condition combines
one simulated recording with the specified
mission and operating settings. Every
policy is evaluated once per condition;
the three cohorts are analyzed separately.}
\label{tab:experimental_cohorts}
\vspace{-4pt}
\footnotesize
\renewcommand{\arraystretch}{1.12}
\begin{tabularx}{\textwidth}{@{}lcccccX@{}}
\toprule
\textbf{Cohort} &
\textbf{Sim. flights} &
\textbf{Duration} &
\textbf{Conditions} &
\textbf{Policies} &
\textbf{Runs} &
\textbf{Varied factors} \\
\midrule

Lake multi-phase &
6 & 120 s & 24 & 21 & 504 &
Mission contract and initial battery;
local perception selection. \\

Depot multi-phase &
4 & 120 s & 16 & 22 & 352 &
Mission contract and initial battery;
local perception selection. \\

Lake dynamic urgency &
6 & 42 s & 72 & 9 & 648 &
Request urgency, modeled connectivity,
and battery; local/remote model routing. \\

\bottomrule
\end{tabularx}
\vspace{-8pt}
\end{table*}
The second study reuses six 42-second Lake
recordings covering three route families
and two focal sectors. 
These recordings use a separate wide-area Lake
layout containing 24 people across eight sectors
(Fig.~\ref{fig:workloads}(a,d)). Each local pass targets three
people in one focal sector; detections of
non-focal people neither satisfy target coverage
nor count as false findings. Each recording
contains a 36-second sensing pass, followed
by an execution and reporting allowance.
The mission requires confirmation and
reporting of at least two of three focal
people using two observations separated
by at least two seconds, with reports
due by 44 seconds. Global limits permit
at most two false detection components,
two incorrect reports, and two duplicate
reports.

The semantic sensing objective remains
fixed; changing urgency modifies the
execution allowance for newly released
perception requests. We consider three
schedules: a constant eight-second
allowance, a reduction from eight to
two seconds at mission time 12 seconds,
and a reduction followed by restoration
to eight seconds at 24 seconds.

Table~\ref{tab:experimental_cohorts}
summarizes the evaluation cohorts.
The multi-phase studies cross each
recording with two mission contracts
and two initial battery conditions.
The dynamic study crosses six recordings
with three urgency schedules, two
network conditions, and two battery
conditions. These combinations reuse
simulated recordings rather than constituting
independent missions.

\vspace{-14pt}
\subsection{Perception Configurations and Resource Profiles}
\label{sec:experimental_configurations}

The multi-phase studies evaluate nine
perception configurations derived from
four pretrained model checkpoints:
Mask R-CNN with a ResNet-50-FPN v2
backbone~\cite{he2017maskrcnn,torchvisionMaskRCNNV2},
Ultralytics YOLO11s and
YOLO11n~\cite{ultralytics2024yolo11},
and CABiNet~\cite{kumaar2021cabinet}
using a checkpoint trained on
UAVid~\cite{lyu2020uavid,cabinetUavidCheckpoint}.

Mask R-CNN contributes standard,
90-degree rotated, and two-view union
configurations. The YOLO11 segmentation
models contribute full-frame inference
and three tiled YOLO11s variants using
standard, rotated, or gamma-adjusted
inputs. The tiled configurations use
a $2\times2$ grid with overlapping
regions; the gamma-adjusted variant
uses $\gamma=0.5$. CABiNet contributes
a semantic-segmentation configuration.
All outputs are converted to a common
binary person-mask representation.
Thus, the nine configurations represent
distinct inference pipelines rather
than nine independently trained models.

\textbf{Primary resource profiles.}
% Previous wording marked for deletion: \textcolor{red}{The multi-phase cohorts use frozen execution profiles derived from offline timing characterization, with configured inference latencies ranging from 0.05 to 6.91 seconds. The two-view Mask R-CNN union accounts for both inference passes and mask fusion.}
The multi-phase cohorts use frozen profiles derived from development-host CPU timings, with configured inference latencies from 0.05 to 6.91 seconds. The two-view union cost sums separately timed inference passes and mask fusion.
We assume a 30 Wh battery, consistent with
the capacity class of lightweight commercial
UAVs~\cite{djiMini4ProSpecs}, and a constant
120 W flight-power budget, comparable to
experimentally characterized rotary-wing UAV
power consumption~\cite{falkowski2023uavpower}.
Onboard inference is modeled at 15 W,
corresponding to a supported power budget
of embedded AI platforms such as the
NVIDIA Jetson TX2~\cite{nvidiaJetsonTx2}.

Model loading incurs 0.3 seconds and 2 J,
routing incurs 0.05 seconds and 0.75 J
per decision, and tracking incurs 0.1
seconds and 1 J. Initial battery charge
is either 34\% or 65\%, with a required
final reserve of 20\%~\cite{njdep2020suasManual}, leaving approximately 0.72 kJ and 34.2 kJ, respectively, for task execution after accounting for flight energy and reserve. All configurations execute locally
in the primary cohorts.

\textbf{Dynamic resource profiles.}
The dynamic-urgency study uses a separately
configured 120 Wh battery model, consistent
with the capacity of commercial UAV
batteries such as the Autel EVO Max
~\cite{autelEvoMaxBattery}. Its ample-energy
condition begins at 95\% charge, whereas
the constrained condition is configured
to leave 250 J for task execution after
accounting for the flight-energy baseline
and the required 20\% reserve.

We consider two synthetic network regimes.
The favorable link provides 50 Mbps
uplink, 100 Mbps downlink, and 30 ms
round-trip time, while the degraded link
provides 2 Mbps uplink, 10 Mbps downlink,
and 150 ms round-trip time. The dynamic study uses Mask R-CNN,
YOLO11s, and CABiNet, each available
through whole-model local or remote
execution. Its profiles account for
input preparation, communication,
server-side processing, result return,
and onboard energy costs. Remote
service times and network behavior
are modeled; the experiments do not
execute inference on a deployed
remote server. 

For all cohorts, candidate predictions
are prepared on the recorded observation
bank. During replay, only the selected
configuration's output becomes available
after its charged completion time.
The physical trajectory and candidate
predictions remain fixed within each
matched comparison, while execution
delays determine which subsequent
observations can be processed.

\vspace{-13pt}
\subsection{Runtime Strategies and Experimental Controls}
\label{sec:experimental_policies}

The multi-phase cohorts compare fixed
perception configurations with adaptive
reference strategies. Fixed policies
repeatedly select one configuration,
including when its incurred resource
costs violate the mission constraints.
The adaptive pool includes round-robin
and predicted-quality selection,
mission- and phase-survival objectives,
temporal evidence-aware selection,
and coverage-oriented policies with
matched ablations.

The Depot cohort additionally evaluates
an adaptation of the quality-under-latency
scheduling objective from
ApproxDet~\cite{xu2020approxdet}.
The comparator uses the benchmark's
perception pool, quality estimates,
and execution profiles; it is not a
reproduction of the complete ApproxDet
detection and scheduling system.

For the multi-phase studies, quality-aware policies
use frozen, per-configuration logistic recall
estimators trained on 596 captures from separate
simulated Lake development flights. Estimates depend on
12 image statistics extracted from the current
capture, without accessing candidate predictions
or evaluation labels. The dynamic study evaluates nine
policies: fixed local Mask R-CNN,
YOLO11s, and CABiNet; fixed remote
Mask R-CNN; a fastest-feasible
selector; and four adaptive
quality-oriented routers. The
adaptive variants differ in whether
they account for estimated
false-detection risk and whether
they observe updated request
deadlines.

Deadline-aware and deadline-blind
variants share the same actual
mission requirements and operating
conditions. The latter continue
selecting configurations using
the initial request allowance,
allowing the effect of responding
to urgency updates to be examined
through matched comparisons.

The dynamic study instead uses fixed per-model
quality and false-detection estimates calibrated
on 162 earlier Lake captures, disjoint from
evaluation images. Local and remote variants
share the same model-quality estimates. To examine how mission requirements guide execution, we also evaluate six reference policies with and without an intent-driven stopping rule. Within each phase, the rule skips later perception once completed reports contain the contract-required number of distinct runtime labels. It uses the public mission contract and the runtime's own reports.

Within every condition, policies
share the same potential observations,
candidate configurations, resource
profiles, and evaluation rules.
Only public system state and outputs
of completed selected actions may
inform subsequent decisions; future
observations, unselected predictions,
and ground-truth identities remain
inaccessible to the policy.

\vspace{-9pt}
\subsection{Evaluation Metrics and Validation Protocol}
\label{sec:experimental_metrics}

Our primary endpoint is complete
mission fulfillment, defined by the
conjunction of applicable requirements
in Sec.~\ref{sec:design}. We additionally
measure phase completion, target coverage,
repeated-evidence confirmation,
timely unique reports, incorrect
and duplicate findings, request
deadline misses, processed and skipped
observations, and resource consumption.
Ground-truth identities are used for
evaluation rather than runtime policy
selection.

For the multi-phase cohorts, predicted person
masks are matched one-to-one against
capture-time ground-truth masks using an
Intersection over Union (IoU) threshold of 0.3. Evidence must become
available within the applicable phase and
reporting deadlines to satisfy temporal
requirements. Predictions completing after
their deadlines retain their incurred
execution costs and applicable false-component
penalties. We perform four additional benchmark validation analyses using the dynamic study’s recorded observations, predictions, and reference executions; resource-accounted hindsight searches are also applied to the multi-phase cohorts.

\textbf{Contract sensitivity.}
We evaluate alternative coverage,
confirmation, revisit, reporting,
and timing requirements through
contract-specific rescoring and
controlled policy reruns. Rescoring
holds execution histories fixed,
whereas reruns expose the modified
contract to the policy and applicable
processing stages. A single-observation
survey relaxation is evaluated only
through rescoring because the current
reporting pipeline requires at least
two observations.

\textbf{Mission feasibility.}
We compute optimistic physical-
opportunity and candidate-model
bounds to identify conditions
that cannot satisfy the required
evidence criteria. We additionally
search for successful hindsight
execution schedules under the
original resource-accounted runtime.
A successful witness establishes
feasibility under the modeled
conditions, whereas an unsuccessful
bounded search remains inconclusive.

\textbf{Causal execution ablation.}
We compare the original serial,
causal replay with a noncausal
no-skip variant that makes otherwise
blocked captures available.
Configured inference/service latency
is varied by factors of
$0.5$, $1$, $2$, and $4$, while
model predictions remain fixed.
Local compute energy scales with latency under the declared power model.
The no-skip protocol is an
analytical ablation rather than
a realizable parallel-execution
implementation.

\textbf{Failure attribution.}
We examine unmet evidence,
reporting, timing, and resource
requirements for unsuccessful
executions. Counterfactual rescoring
relaxes individual requirements
without changing the execution
history, distinguishing diagnostic
changes in mission scoring from
improvements produced by a policy.

All comparisons report the number
of matched policy--condition
executions, while retaining simulated
recordings as the unit of workload
diversity. The dynamic study uses
two recordings for development and
four for comparative evaluation;
all recordings were previously
available and do not constitute
unseen-map generalization.
The validation analyses reuse
these recordings and do not
introduce additional simulated flights.

\vspace{-9pt}
\section{Evaluation}
\label{sec:results}

We evaluate \sys through two complementary
studies of runtime perception and computation
strategies in mobile robotic missions.
Our evaluation examines how perception quality,
execution timing, and resource conditions
influence mission fulfillment, and how
causal execution and mission-level diagnostics
characterize the effectiveness of runtime
decisions. We first examine the multi-phase
Lake and Depot missions, followed by the
dynamic-urgency study and additional
benchmark validation experiments.
% \textcolor{blue}{We evaluate perception and runtime strategies in the multi-phase Lake and Depot missions and the dynamic-urgency study, followed by benchmark validation.}

\vspace{-9pt}
\subsection{Perception Accuracy and Mission Fulfillment}
\label{sec:results_quality}

\begin{figure*}[t]
    \centering
    % \fbox{\parbox[c][2.0in][c]{0.94\textwidth}{
    %     \centering
    %     \textbf{FIGURE 4 PLACEHOLDER}\\[6pt]
    %     (a) Full-bank target recall vs. mission success\\
    %     (b) Predicted vs. observed target recall\\[6pt]
    %     Nine perception configurations in Depot
    % }}
     \includegraphics[width=0.75\textwidth,trim=0bp 24bp 0bp 19bp,clip]{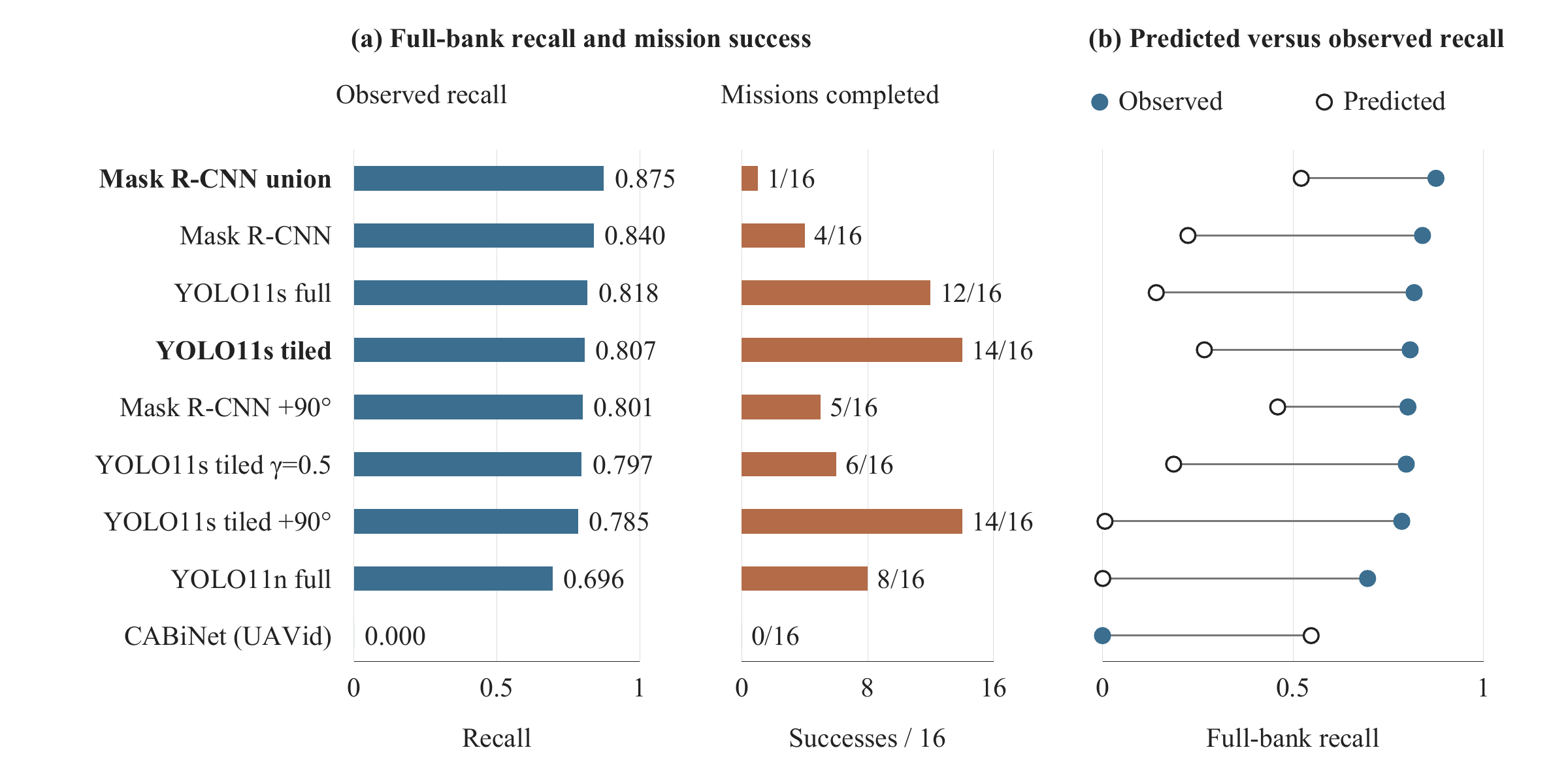}
    \vspace{-4pt}
    \caption{\textbf{Perception quality and mission
    fulfillment across nine configurations in Depot.}
    (a) Full-bank target recall and completed missions
    for fixed perception configurations, highlighting
    the discrepancy between Mask R-CNN union
    (0.875 recall; 1/16 missions) and tiled YOLO11s
    (0.807 recall; 14/16 missions).
    (b) Frozen predicted and observed full-bank
    target recall across configurations, illustrating
    variation in quality-estimation accuracy. {Predicted recall is averaged over captures with visible-target counts as weights for offline analysis.}
    Recall is evaluated over 680 visible target
    observations from 240 potential captures;
    mission success uses 16 resource-accounted
    execution conditions.}
    \label{fig:quality_mission}
\vspace{-12pt}
\end{figure*}

\textbf{Perception quality versus mission success.}
Our first experiment reveals a substantial
discrepancy between frame-level perception
accuracy and mission fulfillment.
Fig.~\ref{fig:quality_mission}(a) compares
full-bank target recall and mission success
across nine fixed perception configurations
in Depot, while Table~\ref{tab:quality_mission}
provides phase-completion and resource-compliance
results for representative configurations.
Full-bank recall is computed over 680
visible target observations across
240 potential RGB captures, whereas
mission success is evaluated over 16
resource-accounted execution conditions.

The two-view Mask R-CNN union achieves
the highest full-bank recall of 0.875,
yet completes only 1 of 16 missions.
In contrast, tiled YOLO11s achieves
a recall of 0.807 and completes
14 of 16 missions. Its rotated tiled
variant likewise completes 14 missions
with a recall of 0.785. The gamma-adjusted tiled variant achieves comparable recall (0.797 versus 0.807) but completes only 6 rather than 14 missions, primarily due to violations of the phase-level false-detection limit. The difference
highlights how perception configurations
with similar individual-frame accuracy
can contribute very differently to
complete robotic missions.

\textbf{Perception quality estimation.}
Fig.~\ref{fig:quality_mission}(b) compares
frozen predicted and observed target recall
across the nine configurations.
The estimates exhibit substantial
configuration-dependent discrepancies.
For example, rotated tiled YOLO11s
achieves an observed recall of 0.785
against a predicted recall of 0.007,
while CABiNet has a predicted recall
of 0.548 and zero observed recall.
These discrepancies characterize
the uncertainty in quality estimates
available to adaptive computation-selection
strategies.

The benchmark further distinguishes
the contributions of perception,
evidence accumulation, reporting,
and resource compliance. The Mask
R-CNN union achieves a mean timely
report recall of 1.000, while
completing 22 of 64 required
mission phases and satisfying the
battery reserve in 8 of 16 conditions. With 6.91 seconds of configured inference latency and two-second capture intervals, Mask R-CNN union processes approximately one in four observation opportunities, limiting the evidence available for phase completion.
In comparison, standard tiled YOLO11s
completes 60 phases and satisfies
the reserve in all 16 conditions.
These results demonstrate how
mission fulfillment emerges from
the combined effects of observation
availability, temporal evidence,
reporting, and resource constraints.

\textbf{Comparison of runtime strategies.}
Across Lake, the evaluated adaptive policies complete 6–13 of 24 missions, while those in Depot complete 2–7 of 16. The coverage-oriented adaptive reference completes 9/24 and 6/16 missions, respectively, whereas the strongest fixed configurations complete 16/24 and 14/16. The ApproxDet-inspired comparator completes 2/16 Depot missions. % These comparisons reveal differences in how runtime strategies utilize available sensing and computation opportunities to satisfy mission requirements.

\textbf{Mission feasibility and runtime effectiveness.}
Resource-accounted, evaluator-guided
search identifies successful execution
schedules for all 24 Lake and
16 Depot conditions. These results
establish achievable mission outcomes
under the evaluated observation
opportunities, perception configurations,
and resource constraints. These feasible execution references quantify the headroom between the evaluated strategies and achievable mission outcomes under the modeled conditions.
\begin{table}[t]
\centering

\caption{\textbf{Perception quality and mission
fulfillment in Depot.}
Full-bank recall and mission success are
reported alongside completed mission phases
and battery-reserve compliance for
representative fixed configurations.
Recall uses 680 visible target observations
across 240 potential captures; mission
outcomes use 16 conditions, each comprising
four required phases.}
\label{tab:quality_mission}
\vspace{-4pt}
\footnotesize
\renewcommand{\arraystretch}{1.10}
\begin{tabularx}{\columnwidth}{@{}Xrrrr@{}}
\toprule
\textbf{Fixed configuration} &
\textbf{Recall} &
\textbf{Success} &
\textbf{Phases} &
\textbf{Reserve} \\
\midrule
Mask R-CNN union &
0.875 & 1/16 & 22/64 & 8/16 \\
YOLO11s, full frame &
0.818 & 12/16 & 58/64 & 16/16 \\
YOLO11s, tiled &
0.807 & 14/16 & 60/64 & 16/16 \\
YOLO11s, tiled + $90^\circ$ &
0.785 & 14/16 & 62/64 & 16/16 \\
\bottomrule
\end{tabularx}
\vspace{-8pt}
\end{table}

\vspace{-9pt}
\subsection{Runtime Adaptation Under Changing Conditions}
\label{sec:results_dynamic}

\begin{figure}[t]
    \centering
    % \fbox{\parbox[c][2.0in][c]{0.94\textwidth}{
    %     \centering
    %     \textbf{FIGURE 5 PLACEHOLDER}\\[6pt]
    %     (a) Mission success, late requests, task energy\\
    %     (b) Success under favorable/degraded connectivity\\
    %     (c) Battery-dependent mission success
    % }}
    \includegraphics[width=\columnwidth,trim=0bp 96bp 0bp 32bp,clip]{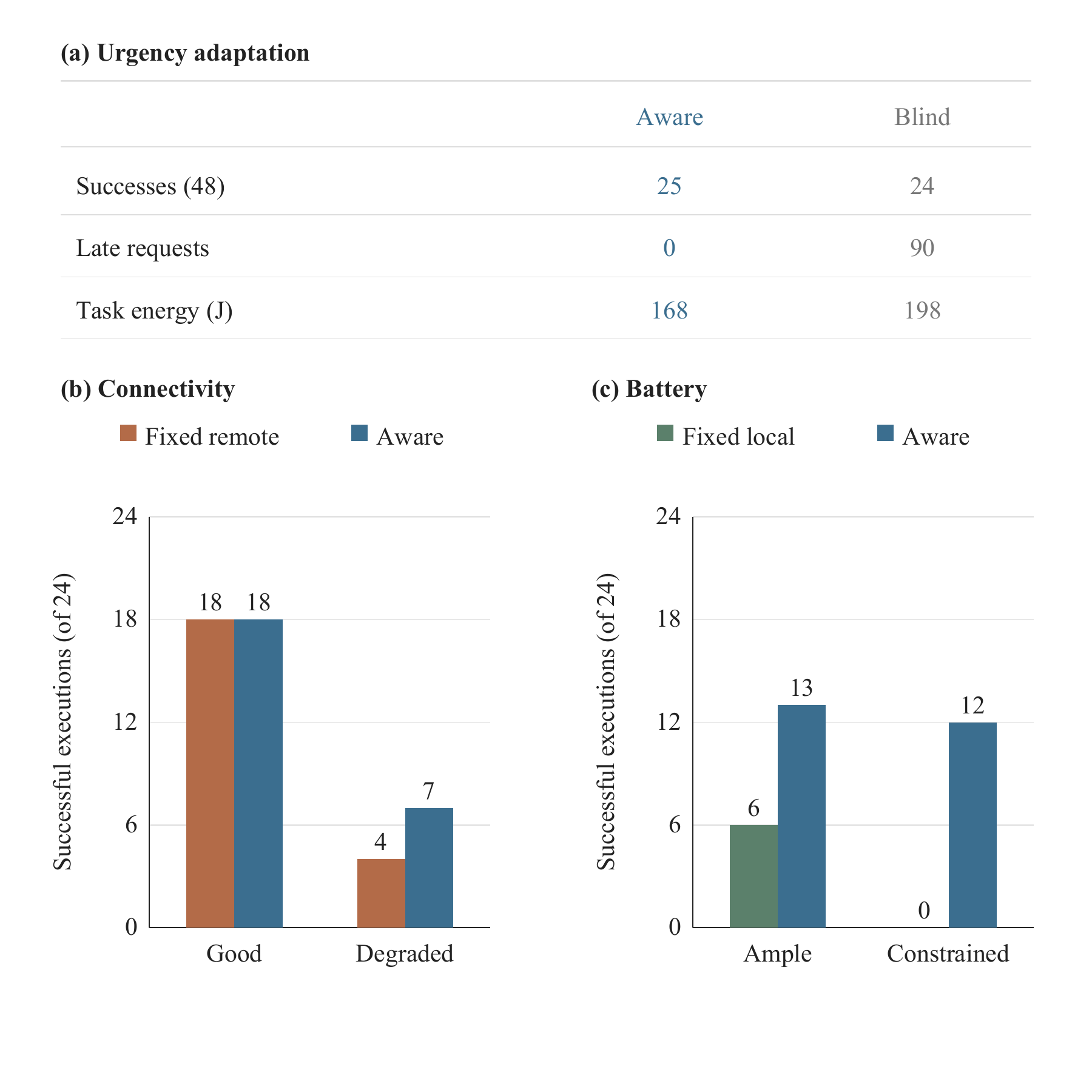}
  \vspace{-4pt}
  \caption{\textbf{Runtime adaptation under changing
urgency and resource conditions.}
(a) Mission success, late perception requests,
and mean onboard task energy for urgency-aware
and urgency-blind quality-and-risk routing across
48 matched conditions.
(b) Mission success under favorable and degraded
connectivity, comparing fixed remote execution
with urgency-aware routing.
(c) Mission success under ample and constrained
battery conditions, comparing fixed local
execution with urgency-aware routing.
Connectivity and battery comparisons use
24 conditions per operating setting.
Energy excludes flight consumption.}
\label{fig:dynamic_adaptation}
\vspace{-11pt}
\end{figure}

We next examine how request urgency,
connectivity, and battery availability
influence runtime decisions and mission
outcomes. The dynamic-urgency study
evaluates nine strategies over
72 conditions; the four-recording
comparative subset contains 48
matched conditions per strategy. 
Fig.~\ref{fig:dynamic_adaptation}
summarizes the effects of these changing
conditions on mission success, execution
timeliness, and resource consumption.

\textbf{Adaptation to changing urgency.}
As shown in Fig.~\ref{fig:dynamic_adaptation}(a),
the urgency-aware quality- and risk-oriented
router completes 25/48 missions,
compared with 24/48 for its
urgency-blind counterpart and 22/48
for fixed remote Mask R-CNN. This
corresponds to three additional
successful conditions, or 6.25
percentage points, relative to
the fixed remote reference.

The influence of urgency awareness
is particularly evident in execution
timeliness. By incorporating updated
request deadlines, the quality- and
risk-oriented router reduces late
perception requests from 90 to zero
across the 48 evaluation conditions.
Its mean modeled onboard task energy
also decreases from 198 J to 168 J,
a reduction of approximately 15\%,
excluding flight energy. The corresponding mission success
changes from 24/48 to 25/48.
The benchmark thus distinguishes
improvements in computational
timeliness and resource efficiency
from their ultimate contribution
to mission evidence and fulfillment.

\textbf{Connectivity-dependent execution.}
Fig.~\ref{fig:dynamic_adaptation}(b)
shows the influence of modeled connectivity
on mission fulfillment. Fixed remote
Mask R-CNN completes 18/24 missions
under favorable connectivity and
4/24 under degraded connectivity.
Under the degraded link, the urgency-aware
quality- and risk-oriented router
completes 7/24 missions. {These results illustrate how network conditions affect mission fulfillment through communication latency and evidence availability.}

% These results demonstrate the
% importance of considering network
% conditions when selecting computation
% configurations for mobile sensing.
% They also illustrate how mission-level
% evaluation captures the combined
% consequences of communication latency,
% result availability, and evidence
% requirements.

\textbf{Energy-constrained execution.}
Fig.~\ref{fig:dynamic_adaptation}(c)
illustrates the impact of battery
availability on mission fulfillment.
Fixed local Mask R-CNN completes
6/24 missions under the ample-energy
condition, compared with 0/24 under
the constrained condition, where
all executions violate the
specified reserve.
This contrast illustrates how
resource obligations influence
mission fulfillment alongside
perception and reporting outcomes. % Together, the urgency, connectivity,
% and energy experiments characterize
% how runtime computing decisions
% interact with changing operating
% conditions, providing complementary
% views of computational efficiency
% and mission-level effectiveness.
\par{\textbf{Embedded hardware sensitivity.} {In replay with Jetson AGX Orin profiles, each adaptive router succeeds in 26/72 conditions with CPU costs and 60/72 with GPU costs. Fixed-local Mask R-CNN succeeds in 0/72 and 60/72, respectively, while fixed-remote Mask R-CNN succeeds in 38/72.} Device and replay capture schedules agree in 10/12 fixed-local runs (Appendix~\ref{app:hardware_sensitivity}). }

\vspace{-9pt}
\subsection{Causal Execution and Observation Availability}
\label{sec:results_causality}

A key feature of \sys is that
computation affects the observation
opportunities available during
physical motion. We evaluate this
relationship through a controlled
comparison between causal execution,
where processing blocks intervening
captures, and a full-capture {counterfactual that retains these
observations. The no-skip protocol permits overlapping inferences as an analytical ablation; it is not an implementation of serial execution.}

\begin{figure}[t]
    \centering
    % \fbox{\parbox[c][2.0in][c]{0.94\textwidth}{
    %     \centering
    %     \textbf{FIGURE 6 PLACEHOLDER}\\[6pt]
    %     (a) Mission success vs. latency multiplier\\
    %     (b) Runtime strategy ranking under\\
    %     causal and full-capture execution
    % }}
    \includegraphics[width=\columnwidth,trim=0bp 43bp 0bp 32bp,clip]{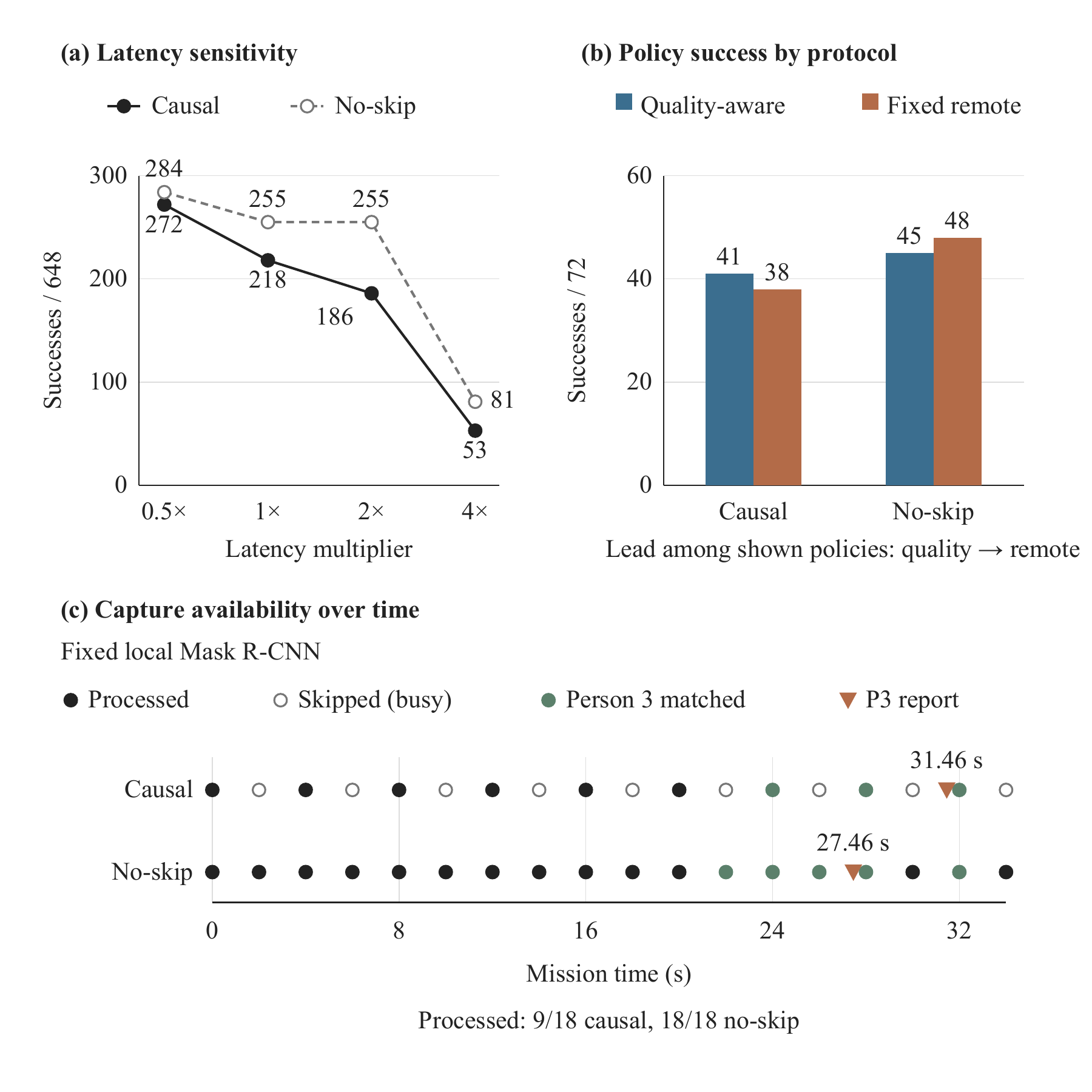}
  \vspace{-4pt}
  \caption{\textbf{Impact of causal execution on
mission outcomes and runtime strategy comparisons.}
(a) Mission successes under causal and no-skip
execution across four inference-latency multipliers,
using 648 matched policy--condition executions
per setting.
(b) Mission successes for urgency-aware
quality routing and fixed remote Mask R-CNN
under causal and no-skip execution.
The relative advantage reverses from
$41/72$ versus $38/72$ under causal
execution to $45/72$ versus $48/72$
under no-skip execution.
{(c) Observation availability in an illustrative fixed-local Mask R-CNN execution. Causal processing admits 9 of 18 captures, compared with 18 under no-skip execution. Markers show capture processing, person-3 matches, and person-level report times.}}
\label{fig:causal_evaluation}
\vspace{-8pt}
\end{figure}

Fig.~\ref{fig:causal_evaluation}(a)
compares mission success under causal
and no-skip execution across four
inference-latency multipliers.
At the original latency, causal
execution yields 218 successful
executions, compared with 255 under
no-skip execution. {The no-skip counterfactual yields 37 additional successes under these modeled conditions.}

% \begin{table}[t]
% \centering
% \caption{\textbf{Impact of causal observation
% availability on mission success.}
% Each protocol evaluates the same
% 648 policy--condition combinations
% per latency multiplier. The
% full-capture counterfactual retains
% observations during blocking computation.}
% \label{tab:causal_ablation}
% \footnotesize
% \renewcommand{\arraystretch}{1.10}
% \begin{tabularx}{\columnwidth}{@{}Xrrr@{}}
% \toprule
% \textbf{Latency} &
% \textbf{Causal} &
% \textbf{Full-capture} &
% \textbf{Difference} \\
% \midrule
% $0.5\times$ & 272 & 284 & +12 \\
% $1\times$   & 218 & 255 & +37 \\
% $2\times$   & 186 & 255 & +69 \\
% $4\times$   & 53  & 81  & +28 \\
% \bottomrule
% \end{tabularx}
% \end{table}

\textbf{Impact on runtime strategy rankings.}
Fig.~\ref{fig:causal_evaluation}(b)
shows how causal observation accounting
changes the relative effectiveness
of runtime strategies.
At the original latency, urgency-aware
quality routing completes 41/72
conditions, compared with 38/72
for fixed remote Mask R-CNN. {Under no-skip execution, the relative ordering of these two strategies reverses, with 45/72 and 48/72 completed missions, respectively.} Across all four adaptive routers, causal execution yields 41–43 successful missions, compared with 38 for fixed remote execution. Under no-skip execution, the corresponding results are 45–48 versus 48, showing that causal observation accounting influences strategy comparisons beyond the representative pair in Fig.~\ref{fig:causal_evaluation} (b).

% This result demonstrates that
% execution-aware observation
% availability can influence
% which runtime strategy appears
% preferable under the same
% mission requirements.

\textbf{Sensitivity to execution latency.}
Doubling configured inference
latency reduces causal mission
success from 218 to 186 executions,
while full-capture success remains
at 255. Among executions that transition from success to failure under causal replay, every one loses at least one observation previously contributing to target confirmation.

{For fixed-local Mask R-CNN, the mean number of processed observations decreases from nine to five as latency doubles.}
These results reveal a direct
connection between computation
latency, available sensing
opportunities, accumulated evidence,
and complete mission fulfillment.

\textbf{Execution-induced observation loss.}
Fig.~\ref{fig:causal_evaluation}(c)
illustrates the effect of blocking
computation on observation availability
during an {illustrative} fixed-local
Mask R-CNN execution. Causal processing
admits nine of 18 potential captures,
whereas no-skip execution processes
all 18. Person-3 reports become available at 31.46 and 27.46 seconds, respectively.
This trace illustrates how computation
latency shapes both the observations
processed and the timing of
accumulated mission evidence.

The evaluation establishes causal
observation accounting as an
important component of benchmarking
runtime computation in mobile
robotic systems. It captures
mission-level consequences that
arise from the interaction between
physical motion and computation.
% Previous Table 4 wording: \textcolor{red}{robotic workloads in simulated or physical environments; runtime inference scheduling or configuration selection.}
% Previous MAVBench and Air Learning View entries: \textcolor{red}{--}.
\begin{table*}[t!]
\centering
\caption{\textbf{Comparison with representative
UAV, embodied, and robotic computing benchmarks.}
Motion: task-level evaluation involving simulated or physical robot motion;
Adapt.: runtime inference scheduling or inference-configuration selection;
View: inference timing affects observation
availability or freshness;
Link: network-aware computation placement;
Energy: energy or power accounting;
Evidence: temporal confirmation and reporting
requirements;
Mission: conjunctive evaluation of evidence,
timing, reporting, and resource requirements.
Checkmarks (\checkmark) denote evaluated
capabilities; dashes (--) indicate features
outside the reported evaluation scope.}
\label{tab:benchmark_comparison}
\vspace{-4pt}

\scriptsize
\setlength{\tabcolsep}{4pt}
\renewcommand{\arraystretch}{1.12}

\begin{tabular*}{\textwidth}{
@{\extracolsep{\fill}}lccccccc@{}}
\toprule

\textbf{Benchmark} &
\textbf{Motion} &
\textbf{Adapt.} &
\textbf{View} &
\textbf{Link} &
\textbf{Energy} &
\textbf{Evidence} &
\textbf{Mission} \\

\midrule

MAVBench~\cite{boroujerdian2018mavbench}
& \checkmark & -- & {\checkmark}
& -- & \checkmark & -- & -- \\

Air Learning~\cite{krishnan2021airlearning}
& \checkmark & -- & {\checkmark}
& -- & \checkmark & -- & -- \\

OpenUAV~\cite{wang2024openuav}
& \checkmark & -- & --
& -- & -- & -- & -- \\

BEDI~\cite{guo2026bedi}
& \checkmark & -- & --
& -- & -- & -- & -- \\

UAVBench~\cite{ferrag2026uavbench}
& -- & -- & --
& -- & -- & -- & -- \\

FloodReasonBench~\cite{bhattacharjya2026floodreasonbench}
& -- & -- & --
& -- & \checkmark & -- & -- \\

\midrule

RobotPerf~\cite{mayoral2024robotperf}
& -- & -- & --
& -- & \checkmark & -- & -- \\

A2Perf~\cite{uchendu2025a2perf}
& \checkmark & -- & --
& -- & \checkmark & -- & -- \\

EmbodiedBench~\cite{yang2025embodiedbench}
& \checkmark & -- & --
& -- & -- & -- & -- \\

LIBERO-Dynamic~\cite{ren2026libero}
& \checkmark & -- & \checkmark
& -- & -- & -- & -- \\

\midrule

Streaming Perception~\cite{li2020towards}
& -- & \checkmark & \checkmark
& -- & -- & -- & -- \\

ApproxDet~\cite{xu2020approxdet}
& -- & \checkmark & --
& -- & -- & -- & -- \\

DEEP~\cite{cong2026decision}
& -- & \checkmark & --
& -- & \checkmark & -- & -- \\

\midrule

\rowcolor{blue!8}
\textbf{\sys (Ours)}
& \checkmark & \checkmark & \checkmark
& \checkmark & \checkmark & \checkmark
& \checkmark \\

\bottomrule
\end{tabular*}
\vspace{-8pt}
\end{table*}
\vspace{-9pt}
\subsection{Mission Feasibility and Evidence-Level Diagnostics}
\label{sec:results_diagnostics}

We further evaluate how \sys
characterizes mission feasibility
and identifies opportunities for
improving runtime strategies.

\textbf{Feasibility characterization.}
We analyze 72 dynamic-urgency conditions using
physical observation bounds, candidate-model
evidence bounds, and resource-accounted hindsight
searches. All 72 satisfy the physical visibility
bound, while 12 are certified infeasible for
the evaluated model pool due to insufficient
repeated confirmations. Reference policies
complete 43 conditions, and hindsight search
identifies successful schedules for eight more;
nine remain unresolved. {These resource-accounted witnesses identify conditions where successful schedules exist under the modeled execution and quantify opportunities beyond the evaluated policies.}

\textbf{Tracing mission evidence.}
Across 216 person–condition instances, the urgency-aware quality-and-risk router obtains matched detections for 172 instances, confirms 139, and generates 129 timely reports. Fixed remote Mask R-CNN achieves 146 matched detections, 114 confirmations, and 112 timely reports. This progression
identifies where evidence is lost between
observation, detection, verification,
and reporting.

\textbf{Requirement-level diagnosis.}
Within the 48-condition comparative subset,
the urgency-aware router completes 25 missions.
Of the remaining 23, all fail the report-recall
requirement, while 18 also lack sufficient
confirmation. Thus, eliminating late requests
does not necessarily provide the evidence
required for mission fulfillment.

\vspace{-9pt}
\subsection{Sensitivity to Mission Requirements}
\label{sec:results_contracts}

% Finally, we examine how mission
% requirements influence evaluation
% outcomes and runtime strategy
% selection. We consider alternative
% contracts that vary confirmation
% thresholds, evidence windows,
% revisit requirements, and reporting
% deadlines, using both contract-specific
% rescoring and controlled policy
% reruns.

% We additionally evaluate 144
% combinations of coverage
% requirements, observation counts,
% false-detection limits, reporting
% obligations, and deadlines.
% These experiments characterize
% the sensitivity of mission-level
% strategy comparisons to the
% specified evaluation criteria. Across 144 rescored contract settings, the
% best adaptive policy selected retrospectively
% for each setting materially outperforms the
% accurate fixed-reference class in 128 settings.
% Here, a material advantage requires
% at least three additional successful
% conditions and no recording favoring
% the comparison class. The accurate
% fixed-reference class exhibits no
% material advantage over adaptive
% routing under this criterion.
% These results characterize the
% robustness of adaptive strategy
% effectiveness across variations
% in mission requirements.

% The benchmark also distinguishes
% changes in mission scoring from
% changes in executed decisions.
% Contract-specific reruns expose
% the modified requirements to
% runtime policies, allowing
% differences in their computation
% selections and mission outcomes
% to be analyzed separately from
% evaluation-only rescoring.

Finally, we examine how mission requirements affect runtime strategy comparisons. We rescore the dynamic-study executions under 144 contract settings varying target coverage, observation requirements, false-detection limits, reporting obligations, and deadlines. For each setting, we compare the retrospectively best adaptive policy against the accurate fixed-reference class (local and remote Mask R-CNN). A material advantage requires at least three additional successful conditions and no recording favoring the comparison class.

Across the 144 settings, the retrospectively selected best adaptive policy achieves a material advantage over both fixed Mask R-CNN references in 128 settings; the reverse advantage is not observed. The grid was specified before these settings were scored, after the five-contract results and diagnostics were available. In separate contract-conditioned reruns {of the original policies}, executed actions changed from the reference schedules in 54 of 1,944 cases, and one mission outcome differed from its fixed-history rescore. {While rescoring examines how requirements affect the assessment of existing executions, the following comparison examines how requirements can guide runtime decisions.} \par{\textbf{Intent-driven execution.} We compare each policy with and without the rule across four contracts and three cost profiles. With the Jetson GPU profile, fixed-local Mask R-CNN uses 28\% and 20\% less estimated task energy under two- and three-observation confirmation, respectively, while completing 60/72 conditions in both cases with and without the rule. Across 5,184 matched pairs of replays, all previously successful executions remain successful. Executing replay-selected schedules on the Jetson reduces mean whole-pass module energy by 7.7\% for two-observation confirmation across six recordings (Appendix~\ref{app:intent_gate}).} {Together, these experiments demonstrate how \sys connects runtime computation decisions to mission evidence and fulfillment under changing physical and resource conditions.}
% \textcolor{blue}{Together, these experiments connect runtime decisions with mission evidence and fulfillment under varying conditions.}

\vspace{-12pt}
\section{Related Work}
\label{sec:related_work}

We now position \sys relative to prior work in
robotic systems benchmarking, embodied task
evaluation, and runtime-aware computation.

\textbf{UAV and robotic systems benchmarks.}
MAVBench~\cite{boroujerdian2018mavbench}
and Air Learning~\cite{krishnan2021airlearning}
evaluate UAV computing systems through
their interactions with physical execution,
task performance, and energy consumption.
The UAV Roofline model~\cite{krishnan2021roofline}
further characterizes the relationship
between sensing, computation, and vehicle
dynamics. RobotPerf~\cite{mayoral2024robotperf}
provides standardized performance evaluation
of robotic computing pipelines across
heterogeneous platforms, while
A2Perf~\cite{uchendu2025a2perf}
evaluates autonomous agents through
task performance, generalization,
reliability, and system resource costs.
System-level characterization of generative
embodied agents~\cite{wan2025generative}
further identifies computation and
communication bottlenecks in
long-horizon embodied tasks.
These studies establish the importance
of relating computational performance
and resource efficiency to the operation
of robotic systems.

\textbf{Embodied task and intent benchmarks.}
OpenUAV~\cite{wang2024openuav}
provides realistic simulated environments
for language-guided UAV navigation.
BEDI~\cite{guo2026bedi} evaluates UAV
agents through perception, decision-making,
and action, using hierarchical metrics
across virtual and real-world scenarios.
UAVBench~\cite{ferrag2026uavbench}
introduces structured UAV mission
scenarios for evaluating agentic reasoning,
while AirCopBench~\cite{zha2026aircopbench}
examines collaborative aerial perception
under challenging sensing conditions. FloodReasonBench~\cite{bhattacharjya2026floodreasonbench}
evaluates VLM reasoning segmentation and
edge execution tradeoffs for flood response.
Beyond UAVs, EmbodiedBench~
\cite{yang2025embodiedbench}
evaluates vision-driven embodied agents
across navigation and manipulation tasks.
LIBERO-Dynamic~\cite{ren2026libero}
incorporates inference latency into
dynamic manipulation, allowing the
simulated environment to evolve while
computation executes. These benchmarks
evaluate complementary aspects of
embodied task execution, instruction
following, perception, and reasoning.

\textbf{Runtime-aware perception and computation.}
Streaming Perception~\cite{li2020towards}
accounts for inference latency and
observations missed during processing,
introducing streaming accuracy as
an online perception metric.
ApproxDet~\cite{xu2020approxdet}
adapts perception configurations
to changing content, latency
requirements, and resource contention,
while Chanakya~\cite{ghosh2023chanakya}
learns runtime configuration decisions
for real-time perception.
DEEP~\cite{cong2026decision}
evaluates perception-computation
allocation through its downstream
decision value under resource budgets.
ACCESS-AV~\cite{10.1145/3771770} adapts localization computation to communication and motion conditions to balance accuracy and energy in autonomous vehicles. MemoGuard~\cite{bhattacharjya2026memoguard} validates retrieved actions against topology, resource, and outcome constraints before memory reuse in communication-limited robot navigation. These approaches demonstrate the
importance of runtime computation
selection and execution timing,
extending evaluation beyond
isolated prediction accuracy.

\textbf{Mission-aware runtime evaluation.}
Table~\ref{tab:benchmark_comparison}
compares representative benchmarks
across robotic execution, runtime
adaptation, and mission-level evaluation.
\sys evaluates runtime computation
strategies over physically simulated
UAV sensing workloads, using recorded
trajectories and common potential
observation opportunities.
Structured mission contracts specify
coverage, repeated confirmation,
fresh revisits, reporting deadlines,
and resource constraints.
Causal, resource-accounted execution
captures how model selection, local
or remote computation, network
conditions, and processing delays
affect the observations and results
available to each mission.
Complete mission fulfillment is
evaluated against these requirements,
with evidence-level diagnostics and
resource-accounted feasibility
witnesses supporting the analysis
of runtime strategy effectiveness.

\vspace{-10pt}
\section{Conclusion and Future Work}
\label{sec:conclusion}

% We presented \sys, a benchmark for evaluating
% intent-driven runtime adaptation in mobile robotic
% missions. By coupling structured mission contracts
% with physically situated sensing and causal,
% resource-accounted execution, \sys connects
% runtime computation decisions to observation
% availability, accumulated evidence, and complete
% mission fulfillment. Our evaluation demonstrates how perception quality, computation-induced observation loss, and resource constraints jointly shape mission fulfillment, including changes in runtime strategy effectiveness and measurable opportunities for improved adaptation. The benchmark further enables requirement-level
% diagnostics and feasibility analysis, providing
% insight into how runtime decisions contribute
% to mission progress and where more effective
% adaptation strategies can improve outcomes.
{We presented \sys, a benchmark for intent-driven runtime adaptation in mobile robotic missions. It couples mission contracts with physically situated sensing and causal, resource-accounted execution to connect computation decisions with observation availability, evidence, and mission fulfillment. Our evaluation shows how perception quality, observation loss, and resource constraints shape mission outcomes and strategy effectiveness. Requirement-level diagnostics and feasibility analysis identify opportunities for improved adaptation.}

% The current evaluation uses Gazebo-based UAV
% workloads with recorded trajectories and modeled
% execution and resource profiles. Future work
% will extend \sys to closed-loop robotic execution
% and measured edge--cloud deployments, enabling
% evaluation of runtime computation under
% coupled sensing, motion, and communication
% dynamics. These extensions will support broader
% investigations of mission-aware adaptation
% across mobile robotic systems.
{The evaluation uses Gazebo-based UAV workloads with recorded trajectories and modeled execution and resource profiles. Future work will extend \sys to closed-loop execution and measured edge--cloud deployments, capturing coupled sensing, motion, and communication dynamics.}

\bibliographystyle{unsrt}
\bibliography{ref_full}

@article{cladera2025air,
  title={Air-ground collaboration for language-specified missions in unknown environments},
  author={Cladera, Fernando and Ravichandran, Zachary and Hughes, Jason and Murali, Varun and Nieto-Granda, Carlos and Hsieh, M Ani and Pappas, George J and Taylor, Camillo J and Kumar, Vijay},
  journal={IEEE Transactions on Field Robotics},
  volume={2},
  pages={626--642},
  year={2025},
  publisher={IEEE}
}

@inproceedings{farahani2026evlm,
  title={{EVLM}: Intent-Driven Edge Vision Language Model for {UAV}-Based Power Line Inspection},
  author={Farahani, Reza and Azimi, Zoha and Murturi, Ilir and Goknil, Arda and Sen, Sagar and Timmerer, Christian and Dustdar, Schahram},
  booktitle={2026 IEEE International Conference on Edge Computing and Communications (EDGE)},
  pages={32--42},
  year={2026},
  organization={IEEE}
}

@article{10.1145/3771770,
author = {Bhattacharjya, Rajat and Sarkar, Arnab and Kool, Ish and Baidya, Sabur and Dutt, Nikil},
title = {ACCESS-AV: Adaptive Communication-Computation Codesign for Sustainable Autonomous Vehicle Localization in Smart Factories},
year = {2026},
issue_date = {January 2026},
publisher = {Association for Computing Machinery},
address = {New York, NY, USA},
volume = {25},
number = {1},
issn = {1539-9087},
howpublished = {\url{https://doi.org/10.1145/3771770}},
doi = {10.1145/3771770},
journal = {ACM Trans. Embed. Comput. Syst.},
month = jan,
articleno = {13},
numpages = {24}
}

@inproceedings{koenig2004design,
  title={Design and use paradigms for gazebo, an open-source multi-robot simulator},
  author={Koenig, Nathan and Howard, Andrew},
  booktitle={2004 IEEE/RSJ international conference on intelligent robots and systems (IROS)(IEEE Cat. No. 04CH37566)},
  volume={3},
  pages={2149--2154},
  year={2004},
  organization={Ieee}
}

@inproceedings{mayoral2024robotperf,
  author = {Mayoral-Vilches, V{\'i}ctor and Jabbour, Jason
    and Hsiao, Yu-Shun and Wan, Zishen
    and Crespo-{\'A}lvarez, Marti{\~n}o
    and Stewart, Matthew and Reina-Mu{\~n}oz, Juan Manuel
    and Nagras, Prateek and Vikhe, Gaurav
    and Bakhshalipour, Mohammad and Pinzger, Martin
    and Rass, Stefan and Panigrahi, Smruti
    and Corradi, Giulio and Roy, Niladri
    and Gibbons, Phillip B. and Neuman, Sabrina M.
    and Plancher, Brian and Reddi, Vijay Janapa},
  title = {{RobotPerf}: An Open-Source, Vendor-Agnostic,
    Benchmarking Suite for Evaluating Robotics
    Computing System Performance},
  booktitle = {2024 IEEE International Conference
    on Robotics and Automation (ICRA)},
  pages = {8288--8297},
  year = {2024},
  doi = {10.1109/ICRA57147.2024.10610841}
}

@inproceedings{uchendu2025a2perf,
  author = {Uchendu, Ikechukwu and Jabbour, Jason
    and Van den Berghe, Korneel and Runevic, Joel
    and Stewart, Matthew and Ma, Jeffrey
    and Krishnan, Srivatsan and Gur, Izzeddin
    and Huang, Austin and Bishop, Colton
    and Bailey, Paige and Jiang, Joe Wenjie
    and Songhori, Ebrahim M. and Guadarrama, Sergio
    and Tan, Jie and Terry, Jordan K.
    and Faust, Aleksandra and Reddi, Vijay Janapa},
  title = {{A2Perf}: Benchmarking Autonomous Agents
    End-to-End in Realistic Domains},
  booktitle = {ICML Workshop on Championing Open-source
    Development in Machine Learning (CODEML)},
  year = {2025},
  howpublished = {\url{https://openreview.net/forum?id=ETrYbzKjat}
}}

@inproceedings{wan2025generative,
  author = {Wan, Zishen and Qian, Jiayi
    and Du, Yuhang and Jabbour, Jason
    and Du, Yilun and Zhao, Yang
    and Raychowdhury, Arijit and Krishna, Tushar
    and Reddi, Vijay Janapa},
  title = {Generative {AI} in Embodied Systems:
    System-Level Analysis of Performance,
    Efficiency and Scalability},
  booktitle = {2025 IEEE International Symposium
    on Performance Analysis of Systems and
    Software (ISPASS)},
  pages = {26--37},
  year = {2025},
  doi = {10.1109/ISPASS64960.2025.00013}
}

@article{azimi2026cloud,
  title={Cloud, Edge, or Split? Profiling Onboard and Split Vision-Language Model Deployment for Drone AI},
  author={Azimi, Zoha and Farahani, Reza and Dustdar, Schahram and Timmerer, Christian},
  journal={arXiv preprint arXiv:2609.25415},
  year={2026}
}

@article{bhattacharjya2026freshlatent,
  title={FRESHLATENT: Channel-Aware Latent Adaptation for Resource-Constrained Embodied VLM Perception},
  author={Bhattacharjya, Rajat and Kim, Minwoo and Sarkar, Arnab and Das, Tamoghno and Wu, Sing-Yao and Bozorgzadeh, Eli and Levorato, Marco and Dutt, Nikil},
  journal={arXiv preprint arXiv:2609.30629},
  year={2026}
}

@article{ren2026libero,
  title={{LIBERO-Dynamic}: Benchmarking Robotic Manipulation in Dynamic Environments},
  author={Ren, Shaojie and Sun, Jingwen and Liu, Zezhi and Chen, Hongjin and Jin, Xin and Zou, Wei and Chen, Zhibo},
  journal={arXiv preprint arXiv:2610.24579},
  year={2026}
}

@article{cong2026decision,
  title={The Decision Value of Perception Compute},
  author={Cong, Hoang Pham and Luong, Ho Viet Duc},
  journal={arXiv preprint arXiv:2609.35910},
  year={2026}
}

@inproceedings{boroujerdian2018mavbench,
  title={{MAVBench}: Micro aerial vehicle benchmarking},
  author={Boroujerdian, Behzad and Genc, Hasan and Krishnan, Srivatsan and Cui, Wenzhi and Faust, Aleksandra and Reddi, Vijay},
  booktitle={2018 51st annual IEEE/ACM international symposium on microarchitecture (MICRO)},
  pages={894--907},
  year={2018},
  organization={IEEE}
}

@inproceedings{mcallister2022control,
  title={Control-aware prediction objectives for autonomous driving},
  author={McAllister, Rowan and Wulfe, Blake and Mercat, Jean and Ellis, Logan and Levine, Sergey and Gaidon, Adrien},
  booktitle={2022 International Conference on Robotics and Automation (ICRA)},
  year={2022},
  organization={IEEE},
  doi={10.1109/ICRA46639.2022.9811884}
}

@inproceedings{li2020towards,
  title={Towards streaming perception},
  author={Li, Mengtian and Wang, Yu-Xiong and Ramanan, Deva},
  booktitle={European conference on computer vision},
  pages={473--488},
  year={2020},
  organization={Springer}
}

@article{bhattacharjya2025avery,
  title={{AVERY}: Intent-Driven Adaptive {VLM} Split Computing via Embodied Self-Awareness for Efficient Disaster Response Systems},
  author={Bhattacharjya, Rajat and Wu, Sing-Yao and Oh, Hyunwoo and Nam, Chaewon and Koo, Suyeon and Imani, Mohsen and Bozorgzadeh, Elaheh and Dutt, Nikil},
  journal={arXiv preprint arXiv:2511.18151},
  year={2025}
}

@misc{gazeboDepotV6,
  author = {Rafi Abdullah and Louise Poubel},
  title  = {{Depot}},
  howpublished    = {\url{https://fuel.gazebosim.org/1.0/OpenRobotics/models/Depot}},
  note   = {Accessed: November 3, 2026}
}

@misc{gazeboHarmonic,
  author = {{Open Robotics}},
  title  = {{Gazebo Harmonic} Documentation},
  howpublished    = {\url{https://gazebosim.org/docs/harmonic/}},
  note   = {Accessed: November 3, 2026}
}

@misc{gazeboHarmonicTerrainV1,
  author = {Cole Biesemeyer},
  title  = {{Harmonic Terrain}},
  howpublished    = {\url{https://fuel.gazebosim.org/1.0/OpenRobotics/models/Harmonic%20Terrain}},
  note   = {Accessed: November 3, 2026}
}

@misc{gazeboHarmonicTerrainObjectsV1,
  author = {Cole Biesemeyer},
  title  = {{Harmonic Terrain Objects}},
  howpublished    = {\url{https://fuel.gazebosim.org/1.0/OpenRobotics/models/Harmonic%20Terrain%20Objects}},
  note   = {Accessed: November 3, 2026}
}

@misc{gazeboLakeHouseV3,
  author = {Cole Biesemeyer},
  title  = {{Lake House}},
  howpublished    = {\url{https://fuel.gazebosim.org/1.0/OpenRobotics/models/Lake%20House}},
  note   = {Accessed: November 3, 2026}
}

@misc{gazeboCoastWaves2V3,
  author = {{OpenRobotics}},
  title  = {{Coast Waves 2}},
  howpublished    = {\url{https://fuel.gazebosim.org/1.0/OpenRobotics/models/Coast%20Waves%202}},
  note   = {Accessed: November 3, 2026}
}

@inproceedings{he2017maskrcnn,
  author    = {Kaiming He and Georgia Gkioxari
               and Piotr Doll{\'a}r and Ross Girshick},
  title     = {{Mask R-CNN}},
  booktitle = {Proceedings of the IEEE International
               Conference on Computer Vision},
  pages     = {2961--2969},
  year      = {2017},
  doi       = {10.1109/ICCV.2017.322}
}

@misc{torchvisionMaskRCNNV2,
  author = {{TorchVision Contributors}},
  title  = {{TorchVision}: Mask R-CNN with
            ResNet-50-FPN v2},
  howpublished    = {\url{https://docs.pytorch.org/vision/main/models/generated/torchvision.models.detection.maskrcnn_resnet50_fpn_v2.html}},
  note   = {Accessed: November 3, 2026}
}

@misc{ultralytics2024yolo11,
  author = {Glenn Jocher and Jing Qiu},
  title  = {{Ultralytics YOLO11}},
  year   = {2024},
  howpublished = {
    \href{https://docs.ultralytics.com/models/yolo11/}
    {docs.ultralytics.com/models/yolo11}
  },
  note = {Accessed: November 3, 2026}
}

@inproceedings{kumaar2021cabinet,
  author    = {Saumya Kumaar and Ye Lyu and
               Francesco Nex and Michael Ying Yang},
  title     = {{CABiNet}: Efficient Context
               Aggregation Network for Low-Latency
               Semantic Segmentation},
  booktitle = {2021 IEEE International Conference
               on Robotics and Automation},
  pages     = {13517--13524},
  year      = {2021},
  doi       = {10.1109/ICRA48506.2021.9560977}
}

@article{lyu2020uavid,
  author  = {Ye Lyu and George Vosselman and
             Gui-Song Xia and Alper Yilmaz
             and Michael Ying Yang},
  title   = {{UAVid}: A Semantic Segmentation
             Dataset for {UAV} Imagery},
  journal = {ISPRS Journal of Photogrammetry
             and Remote Sensing},
  volume  = {165},
  pages   = {108--119},
  year    = {2020},
  doi     = {10.1016/j.isprsjprs.2020.05.009}
}

@misc{cabinetUavidCheckpoint,
  author = {{dronefreak}},
  title  = {{CABiNet (MobileNetV3-Small) Finetuned on UAVid}},
  howpublished    = {\url{https://huggingface.co/dronefreak/cabinet-mobilenetv3-small-uavid}},
  note   = {Accessed: November 3, 2026}
}

@inproceedings{xu2020approxdet,
  author    = {Ran Xu and Chen-lin Zhang
               and Pengcheng Wang and Jayoung Lee
               and Subrata Mitra and Somali Chaterji
               and Yin Li and Saurabh Bagchi},
  title     = {{ApproxDet}: Content and
               Contention-Aware Approximate
               Object Detection for Mobiles},
  booktitle = {Proceedings of the 18th ACM
               Conference on Embedded
               Networked Sensor Systems},
  year      = {2020},
  publisher = {ACM},
  doi       = {10.1145/3384419.3431159}
}

@manual{natsar2026manual,
  author = {{National Search and Rescue Council}},
  title = {{National Search and Rescue Manual:
            Volume 2---Search and Rescue Operations}},
  organization = {Australian Maritime Safety Authority},
  year = {2026},
  howpublished = {\url{https://www.amsa.gov.au/sites/default/files/2026-03/NATSAR-Manual-February-2026-Volume-2.pdf}},
  note = {Accessed: November 3, 2026}
}

@misc{djiMini4ProSpecs,
  author = {{DJI}},
  title  = {{DJI Mini 4 Pro}: Specifications},
  howpublished    = {\url{https://www.dji.com/mini-4-pro/specs}},
  note   = {Accessed: November 3, 2026}
}

@misc{autelEvoMaxBattery,
  author = {{Autel Robotics}},
  title  = {{EVO Max Smart Flight Battery}},
  howpublished    = {\url{https://shop.autelrobotics.com/products/evo-max-smart-flight-battery}},
  note   = {Accessed: November 3, 2026}
}

@misc{nvidiaJetsonTx2,
  author = {{NVIDIA}},
  title  = {{Jetson TX2 Module}: Technical Specifications},
  howpublished    = {\url{https://developer.nvidia.com/embedded/jetson-tx2}},
  note   = {Accessed: November 3, 2026}
}

@article{falkowski2023uavpower,
  author  = {Krzysztof Falkowski and Micha{\l} Duda},
  title   = {Dynamic Models Identification for Kinematics
             and Energy Consumption of Rotary-Wing UAVs
             during Different Flight States},
  journal = {Sensors},
  volume  = {23},
  number  = {23},
  pages   = {9378},
  year    = {2023},
  doi     = {10.3390/s23239378},
  howpublished     = {\url{https://doi.org/10.3390/s23239378}},
  note    = {Accessed: November 3, 2026}
}

@manual{njdep2020suasManual,
  author = {{New Jersey Department of Environmental Protection}},
  title = {{sUAS Flight Operations Manual}},
  organization = {New Jersey Department of Environmental Protection},
  year = {2020},
 howpublished = {\url{https://www.nj.gov/dep/gis/assets/NJDEP_sUAS_Flight_Operations_Manual.pdf}.
}}

@article{krishnan2021airlearning,
  author = {Srivatsan Krishnan and Behzad Boroujerdian
            and William Fu and Aleksandra Faust
            and Vijay Janapa Reddi},
  title = {{Air Learning: A Deep Reinforcement
            Learning Gym for Autonomous Aerial
            Robot Visual Navigation}},
  journal = {Machine Learning},
  volume = {110},
  pages = {2501--2540},
  year = {2021},
  doi = {10.1007/s10994-021-06006-6}
}

@article{guo2026bedi,
  author = {Mingning Guo and Mengwei Wu and
            Jiarun He and Shaoxian Li and
            Haifeng Li and Chao Tao},
  title = {{BEDI: A Comprehensive Benchmark
            for Evaluating Embodied Agents
            on UAVs}},
  journal = {ISPRS Journal of Photogrammetry
             and Remote Sensing},
  volume = {232},
  pages = {910--936},
  year = {2026},
  doi = {10.1016/j.isprsjprs.2026.01.013}
}

@article{ghosh2023chanakya,
  title={Chanakya: Learning runtime decisions for adaptive real-time perception},
  author={Ghosh, Anurag and Balloli, Vaibhav and Nambi, Akshay and Singh, Aditya and Ganu, Tanuja},
  journal={Advances in Neural Information Processing Systems},
  volume={36},
  pages={55668--55680},
  year={2023}
}

@article{krishnan2021roofline,
  title={Roofline model for uavs: A bottleneck analysis tool for designing compute systems for autonomous drones},
  author={Krishnan, Srivatsan and Wan, Zishen and Bhardwaj, Kshitij and Faust, Aleksandra and Reddi, Vijay Janapa},
  journal={arXiv preprint arXiv:2111.03792},
  year={2021}
}

@article{yang2025embodiedbench,
  title={{EmbodiedBench}: Comprehensive benchmarking multi-modal large language models for vision-driven embodied agents},
  author={Yang, Rui and Chen, Hanyang and Zhang, Junyu and Zhao, Mark and Qian, Cheng and Wang, Kangrui and Wang, Qineng and Koripella, Teja Venkat and Movahedi, Marziyeh and Li, Manling and others},
  journal={arXiv preprint arXiv:2502.09560},
  year={2025}
}

@article{ferrag2026uavbench,
  author = {Mohamed Amine Ferrag and
            Abderrahmane Lakas and
            M{\'e}rouane Debbah},
  title = {{UAVBench: An Open Benchmark
            Dataset for Autonomous and
            Agentic AI UAV Systems via
            LLM-Generated Flight Scenarios}},
  journal = {IEEE Open Journal of
             Vehicular Technology},
  volume = {7},
  pages = {2542--2558},
  year = {2026},
  doi = {10.1109/OJVT.2026.3721777}
}

@article{zha2026aircopbench,
  title={{AirCopBench}: A benchmark for multi-drone collaborative embodied perception and reasoning},
  author={Zha, Jirong and Fan, Yuxuan and Zhang, Tianyu and Chen, Geng and Chen, Yingfeng and Gao, Chen and Chen, Xinlei},
  journal={Proceedings of the AAAI Conference on Artificial Intelligence},
  volume={40},
  number={2},
  pages={1507--1515},
  year={2026},
  doi={10.1609/aaai.v40i2.37126}
}

@misc{wang2024openuav,
  author = {Xiangyu Wang and Donglin Yang and
            Ziqin Wang and Hohin Kwan and
            Jinyu Chen and Wenjun Wu and
            Hongsheng Li and Yue Liao and Si Liu},
  title = {{Towards Realistic UAV Vision-Language
            Navigation: Platform, Benchmark,
            and Methodology}},
  year = {2024},
  eprint = {2410.07087},
  archivePrefix = {arXiv},
  howpublished = {
    \href{https://arxiv.org/abs/2410.07087}
    {arXiv:2410.07087}
  }
}

@inproceedings{kim2026approxbit,
  title={{ApproxBit}: Efficient Video Analytics through Latency-Aware Offloading with Learned Binary Codes},
  author={Kim, Hyunseung and Prasanna, Sheetal and Li, Yin and Chaterji, Somali and Bagchi, Saurabh},
  booktitle={Proceedings of the 2026 ACM/IEEE International Conference on Embedded Artificial Intelligence and Sensing Systems},
  pages={1179--1193},
  year={2026}
}

@article{splitsurvey,
  title={Split computing and early exiting for deep learning applications: Survey and research challenges},
  author={Matsubara, Yoshitomo and Levorato, Marco and Restuccia, Francesco},
  journal={ACM Computing Surveys},
  volume={55},
  number={5},
  pages={1--30},
  year={2022},
  publisher={ACM, New York, NY}
}

@inproceedings{10.1145/3774906.3800471,
author = {Xia, Jingfei and He, Yuze and Bian, Chen and Yan, Zhenyu and Xing, Guoliang},
title = {{EventEye}: Towards High-Frequency Perception Enhancement for Autonomous Vehicles Using Infrastructure-Mounted Event Cameras},
year = {2026},
isbn = {9798400723094},
publisher = {Association for Computing Machinery},
address = {New York, NY, USA},
url = {https://doi.org/10.1145/3774906.3800471},
doi = {10.1145/3774906.3800471},
booktitle = {Proceedings of the 2026 ACM/IEEE International Conference on Embedded Artificial Intelligence and Sensing Systems},
pages = {803–817},
numpages = {15},
location = {Saint Malo, France},
series = {SenSys '26}
}

@inproceedings{soyyigit2025mural,
  title={{MURAL}: A multi-resolution anytime framework for {LiDAR} object detection deep neural networks},
  author={Soyyigit, Ahmet and Yao, Shuochao and Yun, Heechul},
  booktitle={2025 IEEE 31st International Conference on Embedded and Real-Time Computing Systems and Applications (RTCSA)},
  pages={33--43},
  year={2025},
  organization={IEEE}
}

@article{chen2026controller,
  title={Controller-aware Perception Neural Network Pruning for CPS Safety Optimization},
  author={Chen, Jingxian and Song, Yukai and Xu, Shengjie and Kessler, Kyle and Ganguli, Prateek and Gupta, Arkaprava and Ghosh, Bineet and Duggirala, Parasara Sridhar and Chen, Tianlong and Chakraborty, Samarjit and others},
  journal={ACM Transactions on Cyber-Physical Systems},
  year={2026},
  publisher={ACM, New York, NY}
}

@inproceedings{weng2026multi,
  title={Multi-Resolution End-to-End Deep Neural Network for Optimizing Latency-Accuracy Tradeoff in Autonomous Driving},
  author={Weng, Qitao and Yun, Heechul},
  booktitle={2026 ACM/IEEE 17th International Conference on Cyber-Physical Systems (ICCPS)},
  pages={12--18},
  year={2026},
  organization={IEEE}
}

@inproceedings{10.1145/3715014.3722081,
author = {Zhao, Minghui and Xia, Junxi and Hou, Kaiyuan and Liu, Yanchen and Xia, Stephen and Jiang, Xiaofan},
title = {{FlexiFly}: Interfacing the Physical World with Foundation Models Empowered by Reconfigurable Drone Systems},
year = {2025},
isbn = {9798400714795},
publisher = {Association for Computing Machinery},
address = {New York, NY, USA},
url = {https://doi.org/10.1145/3715014.3722081},
doi = {10.1145/3715014.3722081},
booktitle = {Proceedings of the 23rd ACM Conference on Embedded Networked Sensor Systems},
pages = {463–476},
numpages = {14},
location = {Irvine, CA, USA},
series = {SenSys '25}
}

@article{bhattacharjya2026floodreasonbench,
  title={{FloodReasonBench}: Benchmarking {VLM} Reasoning Segmentation for Embodied Flood Response at the Edge},
  author={Bhattacharjya, Rajat and Jung, Yoomee and Kim, Minwoo and Wu, Sing-Yao and Bozorgzadeh, Eli and Venkatasubramanian, Nalini and Dutt, Nikil},
  journal={arXiv preprint arXiv:2608.15410},
  year={2026}
}

@inproceedings{bala2024ooda,
  author = {Mihir Bala and Aditya Chanana and Xiangliang Chen and Qifei Dong and Thomas Eiszler and Jingao Xu and Padmanabhan Pillai and Mahadev Satyanarayanan},
  title = {The {OODA} Loop of Cloudlet-Based Autonomous Drones},
  booktitle = {2024 IEEE/ACM Symposium on Edge Computing (SEC)},
  pages = {178--190},
  year = {2024},
  doi = {10.1109/SEC62691.2024.00022}
}

@article{candal2023drone,
  author = {David Candal-Ventureira and Francisco Javier González-Castaño and Felipe Gil-Castiñeira and Pablo Fondo-Ferreiro},
  title = {Is the Edge Really Necessary for Drone Computing Offloading? An Experimental Assessment in Carrier-Grade {5G} Operator Networks},
  journal = {Software: Practice and Experience},
  volume = {53},
  number = {3},
  pages = {579--599},
  year = {2023},
  doi = {10.1002/spe.3161}
}

@inproceedings{yang2022streaming,
  author = {Jinrong Yang and Songtao Liu and Zeming Li and Xiaoping Li and Jian Sun},
  title = {Real-Time Object Detection for Streaming Perception},
  booktitle = {Proceedings of the IEEE/CVF Conference on Computer Vision and Pattern Recognition},
  pages = {5385--5395},
  year = {2022},
  doi = {10.1109/CVPR52688.2022.00531}
}

@article{bhattacharjya2026memoguard,
  author = {Rajat Bhattacharjya and Hyeonjong Ju and Sing-Yao Wu and Eli Bozorgzadeh and Nikil Dutt},
  title = {{MemoGuard}: An Adaptive Runtime for Guarding Against Memory Traps in Communication-Limited Robot Navigation},
  journal = {IEEE Embedded Systems Letters},
  volume = {18},
  number = {5},
  pages = {447--451},
  year = {2026},
  doi = {10.1109/LES.2026.3724555}
}
\clearpage
\appendix
% Supplementary material added for author review.
\balance
\section{Appendix: Reproducibility and Validation Details}
\label{app:reproducibility}

\noindent\textbf{Public artifact:} {\urlstyle{same}\bfseries\url{https://github.com/rajatbh21/EmbodiedIntentBench-artifact}}

This appendix provides additional details for the evaluation in Secs.~4 and~5, including workload validation, mission-contract analyses, causal execution, and Jetson measurements. Several analyses use the dynamic-urgency study, which evaluates adaptation under changing urgency, connectivity, and battery conditions using recorded observations and stored predictions.

The accompanying artifact provides code, inputs, and reference results for this study and the Jetson and intent-driven analyses. The complete benchmark artifact, including the multi-phase workloads and prediction banks, will be released upon acceptance.

\subsection{{Workloads and replay validation}}
{The evaluation comprises multi-phase sensing missions in the Lake and Depot environments and a separate dynamic-urgency study in Lake. The replay validation and contract analyses below focus on the latter study, which varies request urgency and operating conditions over shared recordings.}

{The dynamic-urgency study uses six 42-second Lake recordings with precomputed prediction banks and execution profiles. Its 72 matched conditions combine these recordings with urgency, network, and battery settings. Replay recomputes runtime decisions, evidence, and mission outcomes from stored predictions using modeled execution and network profiles.}

{The accompanying artifact includes the dynamic-study recordings, prediction banks, execution profiles, nine evaluated strategies, reference outcomes, and validation code. The multi-phase recording and prediction bundles are outside its scope; numerical Depot case and frame extracts {support verification of the data underlying Fig.~4.} Validation checks compare regenerated dynamic-study results with reference outputs. Instrumented replay also checked that the evaluated policies accessed predictions only for the selected action and current capture.}

\subsection{{Contract sensitivity protocol}}
{The 144-setting sensitivity analysis rescores the same 648 causal reference histories. These histories correspond to nine evaluated strategies across the 72 operating conditions described above. The grid crosses required coverage of 1, 2, or 3 of the focal people; 1, 2, or 3 observations per person; raw false-component caps of 0, 1, 2, or no cap; required reporting versus sightings-only scoring; and deadlines of 36 or 44 seconds. For $h$ required observations, the minimum observation span is $2(h-1)$ seconds. This grid changes evaluation rules on fixed histories; it does not rerun policies under all 144 contracts. The one-observation survey relaxation is likewise rescoring-only because the reporting pipeline requires at least two observations. Separate contract-specific reruns expose supported contracts to the policy and processing pipeline.}

{For each setting, the adaptive policy with the most successful conditions is selected retrospectively; ties follow the declared policy order. It is compared with each fixed local and fixed remote Mask R-CNN reference. A material advantage requires at least three more successes among the 72 conditions and at least as many successes on every recording. This comparison characterizes the evaluated histories and thresholds.}

\subsection{{Contract-conditioned comparisons}}
{The reference confirmation contract requires reports for at least two of three people, supported by two observations spanning at least two seconds, by 44 seconds. Four variants relax the observation count to one (survey), require three observations spanning four seconds (strict confirmation), require fresh confirmation in both $[0,18)$ and $[18,36)$ seconds (revisit), or move the phase and reporting deadlines to 36 seconds (time-critical). The single-observation survey is distinct from the multi-phase rapid-survey mission, which requires two observations.}

{Table~\ref{tab:appendix_contracts} illustrates how these contracts affect selected reference strategies. The model-union bound pools stored predictions from all candidate models at each capture to provide an optimistic evidence test. For reruns, the denominator includes conditions not certified infeasible by this bound; such conditions can remain unresolved. The survey column is rescored rather than rerun. Across 1,944 eligible contract-conditioned rerun cells, executed actions changed in 54, and one mission outcome differed from fixed-history rescoring. All 12 pairwise rerun ranking reversals were one-condition changes between adaptive variants. }

\begin{center}
\begin{minipage}{\columnwidth}
\captionsetup{hypcap=false}
\captionof{table}{{Selected successes under five contracts. Survey uses rescoring; other columns use reruns. Denominators for reruns exclude only model-union-certified infeasible conditions.}}
\label{tab:appendix_contracts}
\centering

\footnotesize
\setlength{\tabcolsep}{2pt}
\begin{tabular}{lccccc}
\toprule
Strategy & Survey & Confirm & Strict & Revisit & Time \\
\midrule
Local Mask & 12/72 & 12/60 & 8/60 & 0/36 & 12/60 \\
Remote Mask & 38/72 & 38/60 & 34/60 & 12/36 & 38/60 \\
Aware quality+risk & 43/72 & 43/60 & 36/60 & 13/36 & 43/60 \\
\bottomrule
\end{tabular}
\end{minipage}
\end{center}

{The 144-setting grid was specified before its settings were scored, after the five-contract results and descriptive diagnostics were available. The resulting analysis characterizes sensitivity on these inspected recordings.}

\subsection{{Feasibility and causal execution}}
{The feasibility analyses examine whether the recorded observations and candidate predictions can support the mission requirements, while the causal-execution ablation examines how processing delays affect observation availability.}

{The physical-opportunity bounds assume every capture is available at zero latency. The model-union bound additionally allows all candidate models' stored predictions at every capture, with false-component and report caps relaxed. Failure of this optimistic bound certifies infeasibility for the fixed prediction pool under its assumptions; passing it indicates evidence availability under those optimistic assumptions. The ground-truth-aware hindsight search evaluates action scripts through the resource-accounted causal runtime, starting from evaluated policy and constant schedules. It permits up to 600 candidate evaluations and three sweeps per condition. A successful schedule is a feasibility witness, while an unsuccessful bounded search leaves feasibility unresolved.}

{In the no-skip counterfactual, captures are made available even when inference overlaps; completed predictions enter a serial downstream tracker. Concurrent memory is not charged, and the protocol is an analytical ablation rather than a deployable serial implementation. The latency multipliers in Fig.~6 scale configured inference or service latency and the associated local compute energy under the declared power model. Predictions, network transfers, loading costs, and downstream budgets remain fixed.}

\subsection{Embedded hardware sensitivity}
\label{app:hardware_sensitivity}
We profiled Mask R-CNN, YOLO11s, and CABiNet using the benchmark perception pipeline on a Jetson AGX Orin in MODE\_15W (JetPack 6.1, FP32), with four CPU threads or GPU execution. After five warm-up calls, each model processed 24 captures from two recordings; clocks were not pinned. The local replay latency is the empirical 95th percentile.

Power is the sum of the monitored GPU/SOC, CPU/CV, and system 5 V rails; the DDR rail is excluded because it is included in the 5 V rail. The sampler targeted 100 Hz and achieved approximately 60--62 Hz. The replay charges mean monitored power during the profiling block, including idle draw, multiplied by the empirical 95th-percentile latency for each local inference. Idle draw outside inference is not separately charged in this replay; the modeled flight baseline and other costs remain unchanged. These charges are estimates, not directly measured per-call energies.

The dynamic-study replay retains the stored predictions, the original contracts and policies, and modeled remote service and network costs. On the 24 sampled captures, CPU masks for all three models matched the stored masks exactly; on the GPU, Mask R-CNN matched 21/24 exactly (mean IoU 0.9997), while the other two matched 24/24. This is a sample check rather than full-bank parity.

A frame-paced check ran fixed-local Mask R-CNN on the device over six recordings per processor, with non-inference stages emulated and one constant-urgency, favorable-link, ample-battery setting. The resulting capture sets matched replay in 6/6 GPU and 4/6 CPU runs. Device-generated masks were not retained; scoring the observed schedules with cached predictions and an assumed latency is an offline comparison, not an end-to-end device mission test. \subsection{% \textcolor{red}{Contract-aware gating}
 Intent-driven adaptation across mission requirements}\label{app:intent_gate} % \textcolor{red}{For each phase, the gate counts distinct component labels in completed runtime reports whose evidence lies within that phase. It abstains when the count reaches the contract's required number of people. These labels are report-level proxies and need not identify distinct people, so the gate does not itself establish mission fulfillment.}
 To examine how mission requirements can guide perception decisions, we augment the dynamic-study policies with a simple stopping rule and compare them with their original versions. For each mission phase, the stopping rule counts distinct labels in completed runtime reports whose evidence lies within that phase. Once their count reaches the number of people requested by the contract, it skips later perception in that phase. These runtime labels need not identify distinct people; mission success is evaluated separately using target identities and the remaining contract requirements. \par We compare four contracts (confirmation, strict, revisit, and time-critical), six base policies, three cost profiles, and 72 conditions with % \textcolor{red}{the gate on and off}
 the stopping rule enabled and disabled. The 10,368 replays reuse six recordings and stored predictions; remote service and network costs remain modeled. Across 5,184 matched pairs, no success became a failure and three failures became successes. \par% \textcolor{red}{On the Jetson GPU, precomputed fixed-local capture schedules for the confirmation and strict contracts were executed over six recordings each, with and without gating.}
 On the Jetson GPU, fixed-local schedules selected by replay with and without the stopping rule were executed over six recordings for each of the confirmation and strict contracts. Inference was frame-paced, non-inference stages were emulated, and the stopping rule was not executed online on the device. Mean gross module energy during inference fell from 109.1 to 74.4 J and from 107.9 to 84.6 J, respectively; whole-pass module energy fell from 323.7 to 298.9 J and from 323.8 to 305.3 J. Both include idle draw and exclude flight energy. Mission outcomes were scored offline using stored predictions% \textcolor{red}{; the gate was not executed online on the device}
.

\end{document}